\documentclass[10pt,a4paper]{article}
\usepackage[utf8]{inputenc}
\usepackage{amsmath}
\usepackage{amsfonts}
\usepackage{amsthm}
\usepackage{amssymb}
\usepackage{graphicx}
\usepackage{dsfont}
\usepackage{color,xcolor,colortbl}
\usepackage{hyperref} 
\usepackage{authblk}
\usepackage[left=2.5cm,right=2.5cm,top=2cm,bottom=2cm]{geometry}

\usepackage[displaymath, mathlines]{lineno}

\newcommand{\R}{\mathbb{R}}

\definecolor{amaranth}{rgb}{0.9, 0.17, 0.31}
\definecolor{ferngreen}{rgb}{0.31, 0.47, 0.26}

\newcommand{\revised}[1]{\noindent \textcolor{black}{#1}}

\title{Kernel-Based Testing for Single-Cell Differential Analysis}

\author[,1]{A. Ozier-Lafontaine\thanks{To whom correspondence should be addressed: \texttt{anthony.ozier-lafontaine@ec-nantes.fr}, and also \texttt{Bertrand.Michel@ec-nantes.fr}, \texttt{franck.picard@ens-lyon.fr}}}
\author[2]{C. Fourneaux}
\author[2]{G. Durif}
\author[1]{P. Arsenteva}
\author[3,4]{C. Vallot}
\author[2]{O. Gandrillon}
\author[2]{S. Gonin-Giraud}
\author[$*$,1]{B. Michel\thanks{joint last authors}}
\author[$*\dag$,2]{F. Picard}

\affil[1]{Nantes Universit\'e, Centrale Nantes,
Laboratoire de Math\'ematiques Jean Leray, CNRS UMR 6629, F-44000, Nantes, France}
\affil[2]{Laboratory of Biology and Modelling of the Cell, Université de Lyon, Ecole Normale Supérieure de Lyon, CNRS, UMR5239, Université Claude Bernard Lyon 1, Lyon, France}
\affil[3]{CNRS UMR3244, Institut Curie, PSL University, Paris, France.}
\affil[4]{Translational Research Department, Institut Curie, PSL University, Paris, France.}

\DeclareFontFamily{U}{mathx}{\hyphenchar\font45}
\DeclareFontShape{U}{mathx}{m}{n}{
      <5> <6> <7> <8> <9> <10>
      <10.95> <12> <14.4> <17.28> <20.74> <24.88>
      mathx10
      }{}
\DeclareSymbolFont{mathx}{U}{mathx}{m}{n}
\DeclareFontSubstitution{U}{mathx}{m}{n}
\DeclareMathAccent{\widecheck}{0}{mathx}{"71}

\DeclareMathAccent{\wideparen}{0}{mathx}{"75}

\begin{document}
\maketitle

\begin{abstract} 
\revised{Single-cell technologies offer insights into molecular feature distributions, but comparing them poses challenges. We propose a kernel-testing framework for non-linear cell-wise distribution comparison, analyzing gene expression and epigenomic modifications. Our method allows feature-wise and global transcriptome/epigenome comparisons, revealing cell population heterogeneities. Using a classifier based on embedding variability, we identify transitions in cell states, overcoming limitations of traditional single-cell analysis. Applied to single-cell ChIP-Seq data, our approach identifies untreated breast cancer cells with an epigenomic profile resembling persister cells. This demonstrates the effectiveness of kernel testing in uncovering subtle population variations that might be missed by other methods. }
\end{abstract}

\section{Introduction}

Thanks to the convergence of single-cell biology and massive parallel sequencing, it is now possible to create high dimensional molecular portraits of cell populations. This technological breakthrough allows for the measurement of gene expression \cite{macosko_highly_2015,jaitin_massively_2014,zheng_massively_2017}, chromatin states \cite{rotem_single-cell_2015}, and genomic variations \cite{gawad_single-cell_2016} at the single-cell resolution. These advances have resulted in the production of complex high dimensional data and revolutionized our understanding of the complexity of living tissues, both in normal and pathological states. Then, the field of single-cell data science has emerged, and new methodological challenges have arisen to fully exploit the potentialities of single-cell data, among which the statistical comparison of single-cell RNA sequencing (scRNA-Seq) datasets between conditions or tissues. This step has remained a prerequisite in the process to discriminate biological from technical variabilities and to assert meaningful expression differences. While most differential analysis methods primarily focus on expression data, similar methodological challenges have arisen in the comparative analysis of single cell epigenomic datasets, based for example on single cell chromatin accessibility assays (scATAC-Seq \cite{pott_single-cell_2015}) or single cell histone modifications profiling (e.g single-cell ChIP-Seq (scChIP-Seq) \cite{grosselin_high-throughput_2019}, scCUT$\&$Tag \cite{bartosovic_single-cell_2021}). These approaches enable the mapping of chromatin states throughout the genome and their cell-to-cell variations at an unprecedented resolution \cite{shema_single-cell_2019,buenrostro_single-cell_2015} . These single-cell epigenomic assays offer a quantitative perspective on regulatory processes, wherein cellular heterogeneity could drive cancer progression or the development of drug resistance for instance\cite{marsolier_h3k27me3_2022}. The identification of key epigenomic features by differential analysis in disease and complex eco-systems, will be key to understand regulatory principles of gene expression and identify potential drivers of tumor progression. Altogether, comparative analysis of single cell data sets, whatever their type, are an essential component of single cell data science, providing biological insights as well as opening therapeutic perspectives with the identification of biomarkers and therapeutic targets.

Differential Expression Analysis (DEA) is classically addressed by gene-wise two-sample tests designed to detect Differentially Expressed Genes (DEG) \cite{das_differential_2022}. The generalized linear model (GLM) has been a powerful framework for linear parametric testing based on gene-expression summaries \cite{love_moderated_2014,robinson_edger_2010,ritchie_limma_2015}. However, this parametric approach does not fully utilize the entire distribution of gene-expression that characterizes multiple transcriptional states. To achieve the full potential of differential analysis of scRNA-Seq data, DEA has been restated as a comparison between distributions. 
Distributional hypotheses were proposed to capture biologically relevant differences in univariate gene-expressions \cite{korthauer_statistical_2016}. Initially, these tests were performed using Gaussian-based clustering, that was further challenged by distribution-free methods based on ranks or cumulative distribution functions \cite{schefzik_fast_2021,gauthier_distribution-free_2021,tiberi_distinct_2022}. While distribution-free approaches are flexible enough to capture the numerous complex alternatives encountered in DEA, their fully agnostic point of view does not benefit from the significant progress made in modeling scRNA-Seq distributions, which leads to a loss of statistical power. \revised{As a trade-off, we propose a distribution-free test that can still account for certain characteristics of the data, such as a potentially high proportion of zeros.}

Single-cell technologies provide a unique opportunity to obtain a quantitative snapshot of the entire transcriptome, which contains crucial information about between-gene dependencies and underlying regulatory networks and pathways. Therefore, univariate DEA only captures a part of the biological differences and is unable to detect complex global modifications in the joint expression of groups of genes. To fully exploit the complexity of scRNA-Seq data, joint multivariate testing or differential transcriptome analysis should be performed, allowing for cell-wise comparisons. This strategy can be complementary to gene-wise approaches, as the detection of DEG should be interpreted in the context of global differences between conditions. The joint multivariate testing strategy seems also particularly suited to compare epigenomic data since it is well established that chromatin conformation can induce complex dependencies between sites occupancy \cite{margueron_role_2009}. From a distributional perspective, this involves complementing joint distribution-based analyses with analyses based on marginals. Another significant advantage of differential transcriptome analysis is that it can be restricted to targeted GRNs or pathways, allowing for differential network or pathway analyses \cite{mukherjee_distribution-free_2022}. So far, global approaches were mainly developed for differential abundance testing \cite{cao_scdc_2019,dann_differential_2022,buttner_sccoda_2021}, or for the comparison of cell-type compositions. Graph-based methods have been proposed to address differential transcriptome analysis \cite{mukherjee_distribution-free_2022,banerjee_nearest-neighbor_2020}, but they only derive a global $p$-value without any representation or diagnostic tool. 

In recent years, there have been significant advancements in statistical hypothesis testing, alongside the emergence of single-cell technologies. One important breakthrough in hypothesis testing was achieved by Gretton et al. \cite{gretton_kernel_2006}, who combined kernel methods with statistical testing. Kernel methods are widely used in supervised learning \cite{shawe-taylor_kernel_2004} and are based on the concept of embedding data in a feature space, allowing for non-linear data analysis in the input space. Popular dimension reduction techniques, such as tSNE and UMAP  \cite{maaten_visualizing_2008,mcinnes_umap_2018}, also use kernel-based embedding  \cite{van_assel_probabilistic_2022}. The distribution of the embedded data can be described using classical statistics such as means and variances, which can be applied in the feature space. Then the central concept of kernel-based testing is to rely on the Maximum Mean Discrepancy (MMD) test that compares the distance between mean embeddings of two conditions \cite{muandet_kernel_2017}, allowing for non-linear comparison of two gene-expression distributions. Despite the significant potential of kernel-based testing, this approach has not yet been developed in single-cell data science. 

In this work, we propose a new kernel-based framework for the exploration and comparison of single-cell data based on Differential Transcriptome/Epigenome Analysis. Our method relies on  the Kernel Fisher Discriminant Analysis (KFDA) approach introduced by \cite{harchaoui_regularized_2009}. KFDA is a normalized version of the Maximum Mean Discrepancy to account for the variability of the datasets. This results in a test statistic that can be interpreted as the distance between mean embeddings projected onto the kernel-Fisher discriminant axis. Although KFDA was initially introduced as a non-linear classifier \cite{mika_fisher_1999}, it is a great example of how classifiers can be used for hypothesis testing \cite{harchaoui_kernel-based_2013,lopez-paz_revisiting_2018}, and recent developments have demonstrated its optimality \cite{hagrass_spectral_2022}. Here we show that the KFDA-witness function, which is the Fisher discriminant axis \cite{kubler_witness_2022}, can further be used for data exploration of scRNA-Seq and scChIP-Seq data. Our method is available in a package called \texttt{ktest} \footnote{\url{https://github.com/LMJL-Alea/ktest}} available in both R and Python, which offers many visualization tools based on the geometrical concepts from the Fisher Discriminant Analysis (FDA) to aid comparisons. \revised{While originally designed for a two-sample framework, our method can be extended to accommodate multiple group comparisons. Furthermore, we discuss its applicability and extension to more complex experimental designs.} We show the calibration and the power of our method compared with others on simulated \cite{gauthier_distribution-free_2021} and multiple scRNA-Seq datasets \cite{squair_confronting_2021}. Then we illustrate the power of the classification-based testing approach, that identifies sub-populations of cells based on expression and epigenomic data, that would not be detected otherwise. When applied to scRNA-Seq data, \texttt{ktest} reveals the heterogeneity in differentiating cell populations induced to revert toward an undifferentiated phenotype \cite{zreika_evidence_2022}. Our method also uncovers the epigenomic heterogeneity of breast cancer cells, revealing the pre-existence - prior to cancer treatment - of cells epigenomically identical to drug-persister cells, i.e the rare cells that can survive treatment.

As single-cell datasets grow larger and more complex, traditional testing methods may fail to capture subtle variations and accurately identify meaningful  differences in molecular patterns. Here we show that kernel testing emerges as a promising approach to overcome these challenges, offering a robust and flexible framework. Kernel testing techniques are less sensitive to assumptions on data distribution than traditional methods, and can handle complex dependencies within and across cells. This capability is particularly relevant in the context of single-cell data, where inherent noise, sparsity, and heterogeneity pose unique challenges to accurate statistical inference. Overall, kernel testing represents a powerful tool for the differential analysis of single-cell data, enabling to uncover hidden patterns, and gain deeper insights into the intricate heterogeneities of cell populations.

\section{Results}
In the following we denote by $Y_1 = (Y_{1,1},\dots,Y_{1,n_1})$ and $Y_2=(Y_{2,1},\dots,Y_{2,n_2})$ the gene expression measurements of $G$ genes with distributions $\mathbb{P}_1$ and $\mathbb{P}_2$ in conditions 1 and 2 on $n_1$ and $n_2$ cells respectively, $n=n_1+n_2$. In the following, we will derive our method for expression data, but it can be generalized to any single-cell data. Then we suppose that \begin{eqnarray*}
    Y_{i,j} \sim \mathbb{P}_i, \quad i = 1,2 \quad j = 1, \dots n_i.
\end{eqnarray*}
Two-sample testing between distributions consists in challenging the null hypothesis  $H_0:\mathbb{P}_1=\mathbb{P}_2$ by the alternative hypothesis $H_1: \mathbb{P}_1\neq\mathbb{P}_2$. To construct a non-linear test we consider the embeddings of the original data denoted by $\big( \phi(Y_{i,1}),\dots,\phi(Y_{i,n_i}) \big)$ ($i=1,2$), obtained using the feature map $\phi$ that maps the data into the so-called feature space $\mathcal{H}$ that is a reproducing kernel Hilbert space. The kernel provides a measure of the similarity between the observations, that turns out to be the inner product between the embeddings: 
\begin{eqnarray*}
k(Y_{i,j},Y_{i',j'}) = \langle \phi(Y_{i,j}), \phi(Y_{i',j'}) \rangle_{\mathcal{H}}.
\end{eqnarray*}
Thanks to this relation, kernel methods are non-linear for the original data, but linear with respect to the embeddings in the feature space. They provide a non-linear dissimilarity between cells based either on the whole transcriptome or on univariate gene distributions. Kernel-based tests consist in the comparison of kernel mean embeddings of distributions $\mathbb{P}_1$ and $\mathbb{P}_2$ \cite{muandet_kernel_2017}, defined such that:
\begin{eqnarray*}
\forall i \in \{1,2\}, \quad \mu_i = \mathbb{E}_{Y \sim \mathbb{P}_i} \left[ \phi(Y) \right].    
\end{eqnarray*}
The initial contribution to kernel testing involved calculating the distance between kernel mean embeddings with the MMD statistic \cite{gretton_kernel_2012}. However, it is difficult to determine its null distribution, and since the MMD does not account for the variance of embedding, it has recently been show to lack optimality \cite{hagrass_spectral_2022}. By utilizing a Mahalanobis distance to standardize the difference between mean embeddings, we can not only obtain an asymptotic chi-square distribution for the resulting statistic \cite{harchaoui_kernel-based_2013}, but we can also take advantage of the kernel Fisher Discriminant Analysis (KFDA) framework that is typically used for non-linear classification. Therefore, we present two complementary perspectives on the KFDA testing framework: one based on a distance-based construction of the statistic and the other on the kernel FDA, which offers several visualization tools to highlight the main cell-wise differences between the two tested conditions.

\subsection{Testing with a Mahalanobis distance between gene-expression embeddings}

The squared distance between the kernel mean embeddings constitutes the so-called Maximum Mean Discrepancy statistic, such that:
\begin{eqnarray*}
\operatorname{MMD}^2(\mu_1,\mu_2) &=& \|\mu_1 - \mu_2\|^2_{\mathcal{H}} \\
&=&\mathbb{E}_{Y_1 \sim \mathbb{P}_1, Y_1' \sim \mathbb{P}_1} \left[ k(Y_1,Y_1') \right] + \mathbb{E}_{Y_2 \sim \mathbb{P}_2, Y_2' \sim \mathbb{P}_2} \left[ k(Y_2,Y_2') \right] \\
&-& 2 \times \mathbb{E}_{Y_1 \sim \mathbb{P}_1, Y_2 \sim \mathbb{P}_2} \left[ k(Y_1,Y_2) \right].
\end{eqnarray*}
This statistic tests the between-class separation by comparing expected pairwise similarities between and within conditions 1 and 2 \revised{(a full derivation is proposed in the Supplementary Material)}. 
\revised{To account for the residual variability, we introduce the weighted Mahalanobis distance between mean embeddings, 
\begin{eqnarray*}
\operatorname{D}_T^2(\mu_1,\mu_2) = \frac{n_1n_2}{n}\|\Sigma_{W,T}^{-1/2}\left( \mu_1 - \mu_2 \right) \|^2_{\mathcal{H}},
\end{eqnarray*}
where $\Sigma_{W,T}$ contains the first $T$ principal directions of the homogeneous within-group covariance of embeddings defined such as:
\begin{eqnarray*}
\Sigma_W = \frac{n_1}{n} \Sigma_1 + \frac{n_2}{n} \Sigma_2,
\end{eqnarray*}
with
\begin{eqnarray*}
\forall i \in \{1,2\}, \quad \Sigma_i = \mathbb{E}_{Y \sim \mathbb{P}_i}\left [ (\phi(Y) - \mu_i)^{\otimes 2} \right ],
\end{eqnarray*}
the covariance operator within each condition ($\otimes$ stands for the outer product in the feature space). Regularization is indeed necessary to prevent the singularity of $\Sigma_W$. One potential approach is to introduce ridge regularization; however, this leads to a complex distribution of the test statistic under the null hypothesis, with limited interpretability \cite{harchaoui_testing_2008}. An alternative regularization strategy consists in considering $\Sigma_{W,T}$ which involves a kernel-PCA dimension-reduction step  to capture the residual variability of expression data centered by condition.} Then the corresponding regularized statistic is based on the estimated mean embeddings and covariances:
\begin{eqnarray*}
\forall i \in \{1,2\}, \quad \widehat{\mu}_i = \frac{1}{n_i}\sum_{j = 1}^{n_i} \phi(Y_{i,j}),\quad \widehat{\Sigma}_i = \frac{1}{n_i}\sum_{j = 1}^{n_i} (\phi(Y_{i,j}) - \widehat{\mu}_i))^{\otimes 2}.
\end{eqnarray*}
The main computational complexity comes from the eigen-decomposition of $\widehat{\Sigma}_W = ( n_1 \widehat{\Sigma}_1 + n_2 \widehat{\Sigma}_2 ) / n$ which requires  $O(n^3)$ operations and results in the truncated covariance $\widehat{\Sigma}_{W,T}=\sum_{t=1}^T\widehat{\lambda}_t (\widehat{e}_t \otimes \widehat{e}_t)$, where $(\widehat{\lambda}_t)_{t=1:T}$ are the decreasing eigenvalues of $\widehat{\Sigma}_{W,T}$ and $(\widehat{e}_t)_{t=1:T}$ are the associated eigenfunctions referred by extension in the following as principal components. Then the empirical weighted Mahalanobis distance between the two mean-embeddings is :
\begin{eqnarray*}
 \widehat{\operatorname{D}}^2_T(\widehat{\mu}_1,\widehat{\mu}_2) = \frac{n_1n_2}{n}\left \| \widehat{\Sigma}_{W,T}^{-\frac{1}{2}} (\widehat{\mu}_2 - \widehat{\mu}_1) \right \| _{\mathcal{H}}^2.
\end{eqnarray*}
This statistic follows a $\chi^2(T)$ asymptotically under the null hypothesis \cite{harchaoui_regularized_2009}, which resumes to the Hotelling's test in the feature space. Using the asymptotic distribution for testing seems reasonable for scRNA-Seq data for which $n \geq 100$, otherwise, it is possible to test with a permutation procedure for small sample sizes. Our implementation runs in $\sim 5$ minutes for $n \sim 4000$, and the package proposes a sampling-based Nystrom approximation for larger sample sizes \cite{williams_using_2001}. 

\subsection{The kernel Fisher Discriminant analysis, a powerful tool for non-linear DEA}

A major advantage of using the Mahalanobis distance between distributions is that the test statistic can be reinterpreted under the light of a classification problem, thanks to its connection with the Fisher Discriminant Analysis (FDA). This framework induces a powerful cell-wise visualization tool that allows to explore and understand the nature of the differences between transcriptomes. FDA is a linear classification method that consists in finding the linear axis that optimizes the discrimination between the two distributions. Intuitively, a direction is discriminant if the observations projected on it ($i$) do not overlap and ($ii$) are far from each other. Hence the best discriminant axis is found by maximizing the Fisher Discriminant Ratio, that models a trade-off between minimizing the overlap while maximizing the distance between the means of the two groups. By finding this linear axis in the feature space to classify the embeddings, we obtain a non-linear function that makes the two distributions linearly separable. Thus, in the feature space we denote by $h^\star_T$ the optimal axis that maximizes the truncated Fisher Discriminant Ratio :
\begin{eqnarray*}
h^\star_T = \underset{h \in \mathcal{H}}{n \: \operatorname{argmax}} \frac{\left \langle h , \Sigma_B h \right \rangle_{\mathcal{H}}}{\left \langle h , \Sigma_{W,T} h \right \rangle_{\mathcal{H}}}.
\end{eqnarray*}
where $\Sigma_B$ is the between-group covariance capturing the part of the variance of the embeddings due to the difference between the two groups : 
\begin{eqnarray*}
\Sigma_B = \frac{n_1n_2}{n^2} (\mu_1 - \mu_2)^{\otimes 2}.
\end{eqnarray*}
The numerator of the Fisher Discriminant Ratio captures the distance between the two mean embeddings on a given direction, to be maximized, and the denominator captures the variability of the embeddings projected on this direction, standing for a measure of the overlap, to be minimized. The discriminant axis $h^\star_T$ can be found in closed form from an analytical reasoning. The Mahalanobis distance then appears to be the maximal value of the ratio, which is the distance between the mean embeddings projected on $h^\star_T$ :
\begin{eqnarray*}
\operatorname{D}^2_T = n \frac{\left \langle h_T^\star , \Sigma_B h_T^\star \right \rangle_{\mathcal{H}}}{\left \langle h_T^\star , \Sigma_{W,T} h_T^\star \right \rangle_{\mathcal{H}}} = \frac{n_1n_2}{n}\|\Sigma_{W,T}^{-1/2}\left( \mu_1 - \mu_2 \right) \|^2_{\mathcal{H}},
\end{eqnarray*}
By relying on both the within-group and the between-group covariances, the FDA approach encompasses the total variability of the embeddings. We can interpret the projection of the embeddings on $h^\star_T$ in terms of similarity between the two groups. The extreme values of projected embeddings on the discriminant axis correspond to cells that contain the most significant information for distinguishing between conditions. Conversely, the central values of projected embeddings correspond to cells that do not contribute to the discrimination and hold less informative value. We will propose an illustration to show how this representation can be used to identify outliers or sub-populations.

Then non-linear testing turns out to be very powerful to detect complex alternatives, like the ones proposed in the context of distribution-based DEA \cite{korthauer_statistical_2016}. We illustrate the discriminant axis by representing the four standard alternative hypotheses: differential mean (DE), differential proportions (DP), differential modality (DM) and differential both proportion and modality (DB)  \cite{korthauer_statistical_2016}. The DE, DP and DM alternatives are somehow easy to discriminate even with summary statistics because the distributions have different means, projecting the embeddings on the discriminant axis easily discriminates the two conditions. On the contrary, the DB alternative is the most difficult alternative to detect with many DEA approaches, because the two conditions share the same mean expression \cite{gauthier_distribution-free_2021}. The discriminant axis acts as a powerful non-linear transformation of the expression data to make the two distributions easily separable (Fig. \ref{fig:data_ccdf}). \revised{For the sake of simplicity, we presented our method in the two-sample setting, but we also propose a generalization to multiple groups comparisons provided in the Supplementary Material.}

\subsection{Kernel Choice \label{Section:kernel_choice}}

The design of appropriate kernels is an active field of research \cite{bach_multiple_2004, schrab_mmd_2022}. In kernel-based testing, choosing an appropriate kernel has many objectives like capturing important data characteristics and showing sufficient power to distinguish between different alternatives. To this extent, the conclusions drawn in the feature space from the mean embeddings should apply to the initial distributions. In other words, it should be equivalent to test $\mu_1=\mu_2$ for $\mathbb{P}_1=\mathbb{P}_2$ which is not true in general. However, both are equivalent for a particular class of kernels called universal kernels, which has lead to theoretical and computational developments \cite{simon-gabriel_kernel_2018,gretton_optimal_2012,schrab_mmd_2022}. Fortunately, the Gaussian kernel fulfills this universality property. For two cells $\{(i,j),(i',j')\}$ and genes $g=1,\hdots, G$, our developments will be based on $k_{\operatorname{Gauss}}$ defined such that :
\begin{eqnarray*}
k_{\operatorname{Gauss}}(Y_{i,j},Y_{i',j'}) =  \exp \left(-\frac{1}{2 \sigma^2} \sum_{g=1}^{G}(Y^g_{i,j}-Y^g_{i',j'})^2 \right).
\end{eqnarray*}
This kernel can be used in both multivariate and univariate contexts. Once the Gaussian kernel has been chosen, the remaining question concerns the calibration of its bandwidth $\sigma$, which is done using the median heuristic \revised{that consists in choosing $\widehat{\sigma}^2 = \operatorname{median}( \sum_{g}( Y^g_{i,j}-Y^g_{i',j'})^2, (i, i') \in \{1,2\}^2, j \in \{1, \hdots, n_i\}, j' \in \{1, \hdots, n_{i'}\})$} \cite{gretton_optimal_2012,schrab_mmd_2022,garreau_large_2018}. \revised{Depending on the sequencing technology \cite{svensson_droplet_2020}, scRNA-Seq data may contain a fraction of zeros (especially for non-UMI data like Smart-Seq, for instance), which could impact the calibration of the kernel's bandwidth if not properly considered. Therefore, we propose a two-compartment kernel based on probability product kernels \cite{jebara_probability_2004}. Let $\pi_i$ represent the proportion of zeros in condition $i$, and $f_{\mu,\sigma}$ denote the Gaussian probability function. We introduce a zero-inflated Gaussian kernel (details in the Methods Section):
\begin{eqnarray*}
k_{\text{ZI-Gauss}}(Y_{i,j},Y_{i',j'}) &=& \pi_i \pi_{i'} + \pi_i(1-\pi_{i'})f_{\mu_{i'},\sigma}(0) + (1-\pi_i)\pi_{i'}f_{\mu_i,\sigma}(0) \\
&+&(1-\pi_i)(1-\pi_{i'}) k_{\operatorname{Gauss}}(Y_{i,j},Y_{i',j'}), 
\end{eqnarray*}
so that the bandwidth is calibrated on non-zero entries only. Finally, in our method comparisons, we will explore the \texttt{ktest} framework with a linear kernel to highlight the advantages of non-linearity. For this illustration, we consider the standard scalar product:
\begin{eqnarray*}
k_{\operatorname{linear}}(Y_{i,j},Y_{i',j'}) = \sum_{g=1}^G Y_{i,j}^g \times Y_{i',j'}^g.
\end{eqnarray*}}

\subsection{Kernel testing is calibrated and powerful on simulated data}

\begin{figure}
\begin{center}
\includegraphics[scale=0.25]{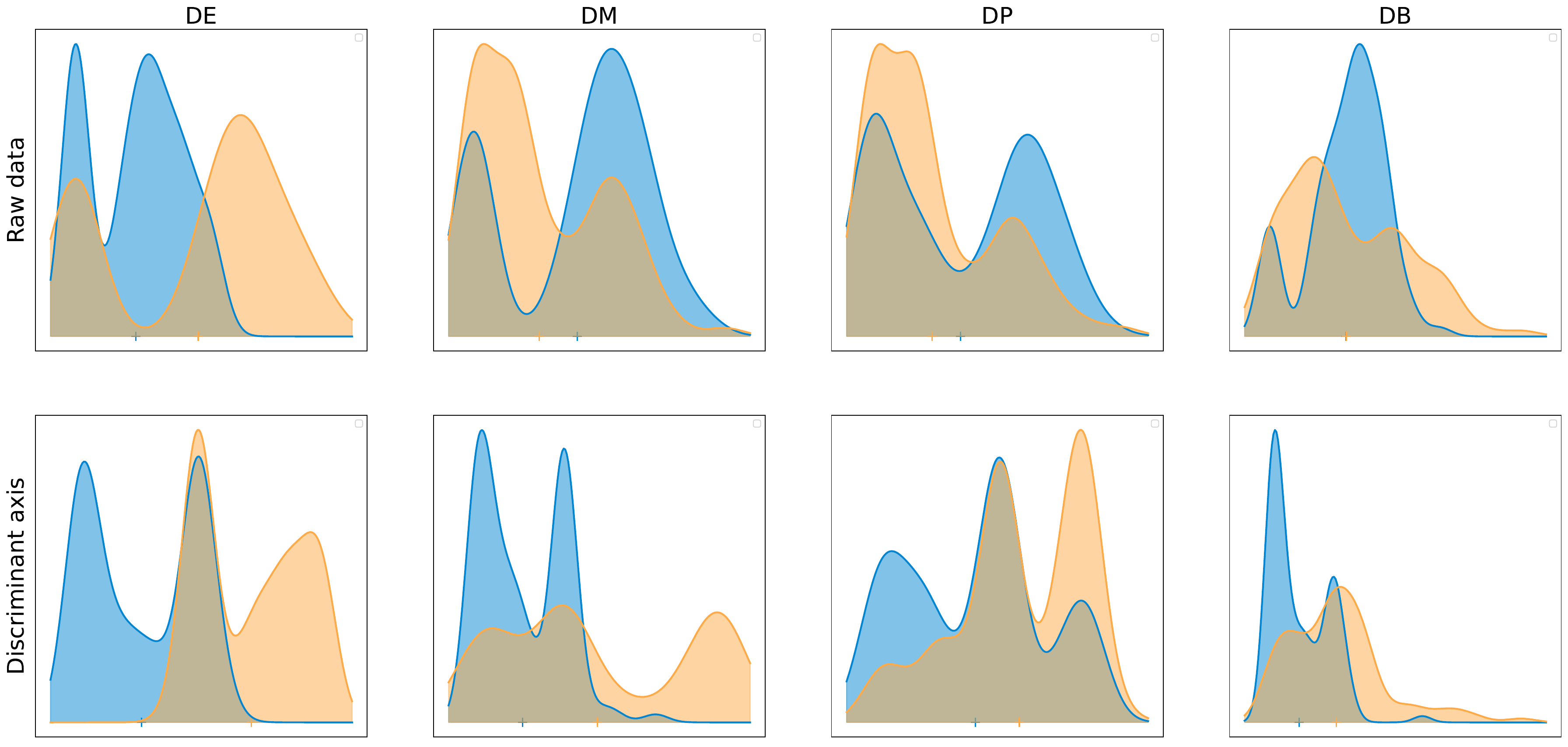}
\end{center}
   \caption{Top : Examples of distributions of the simulated data, DE : classical difference in expression, DM : difference in modalities, DP : difference in proportions, DB : difference in both modalities and proportions with equal means. Bottom : projection of cells on the discriminant axis ($T=4
$) for each alternative. The non-linear transform allows the separation of distributions on the discriminant axis.}
\label{fig:data_ccdf}
\end{figure}

\begin{figure}
\begin{center}
\includegraphics[scale=0.25]{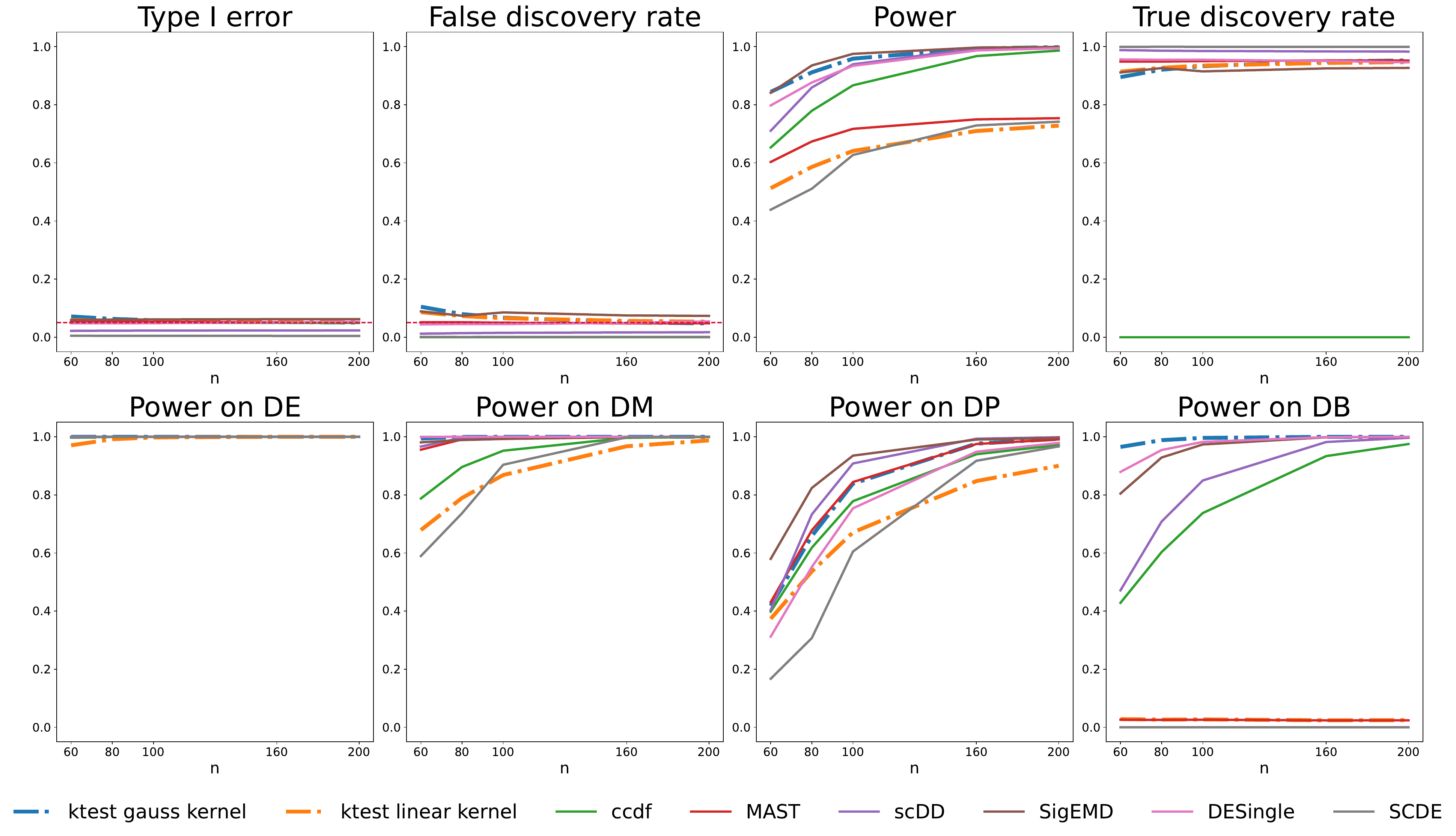}
\end{center}
   \caption{Comparison of DEA methods with respect to type-I errors and power. Top: Type-I errors are computed on raw $p$-values under $H_0$. False discovery Rate computed on Benjamini-Hochberg adjusted $p$-values. Power computed on raw $p$-values under $H_1$. True Discovery Rate computed on Benjamini-Hochberg adjusted $p$-values. Simulated data consists of $100$ cells, $10000$ genes ($1000$ DE, $9000$ non-DE). Alternatives are simulated using DE : classical difference in expression ($250$ genes), DM : difference in modalities ($250$ genes), DP : difference in proportions ($250$ genes), DB : difference in both modalities and proportions with equal means ($250$ genes). Error rates are computed over $500$ replicates. \revised{The truncation parameter is set to $T=4$ for the Gauss-kernel.}} 
\label{fig:performances_ccdf}
\end{figure}

Simulations are required to compare the empirical performance of DE methods on controlled designs, to check their type-I error control and compare their power on targeted alternatives. \revised{We challenged our kernel-based test with six standard DEA methods (Table \ref{tab:methods_simu}) on mixtures of zero-inflated negative binomial data reproducing the DE, DM, DP and DB alternatives \cite{gauthier_distribution-free_2021} (as detailed in Material and Methods).} \revised{Kernel testing was performed on the raw data using the Gauss and ZI-Gauss kernels, but we also considered the linear kernel (scalar product) to illustrate the interest of a non-linear method.} The type-I errors of the kernel test are controlled at the nominal levels $\alpha=5\%$ and the performance increases with $n$ (the asymptotic regime of the test is reached for $n\geq 100$). The Gauss-kernel test is the best method for detecting the DB alternative, considered as the most difficult to detect, and it outperforms every other method in terms of global power excepted SigEMD. This gain in power can be explained by the non-linear nature of our method: despite the equality of means, the kernel-based transform of the data onto the discriminant axis allows a clear separation between distributions (Fig. \ref{fig:data_ccdf}). \revised{This is well illustrated by the global lack of power of the test based on the linear kernel (especially on the DB alternative).} The Gaussian kernel shows its worst performances on the DP alternative, which is the only alternative for which all the values are covered by both conditions with different proportions. It shows that our method is particularly sensitive to alternatives where some values are occupied by one condition only (Fig. \ref{fig:performances_ccdf}). Note that the ZI-Gauss kernel did not improve the global performance, which indicates that the Gaussian kernel-based test is robust to zero inflation. This could also be due to the equality of the zero-inflation proportions between conditions. Finally, results on log-normalized data are similar. \revised{We also checked that the median heuristic was a reasonable choice for the bandwidth parameter (Fig \ref{fig:bandwidth}), as it established a good type-I/power trade-off. Note that when the bandwidth of the Gaussian kernel increases, the truncation parameter should be calibrated accordingly to reach the same type-I/power performance.}

\subsection{Challenging DEA methods on experimental scRNA-Seq data}

\begin{figure}
\begin{center}
\includegraphics[scale=0.35]{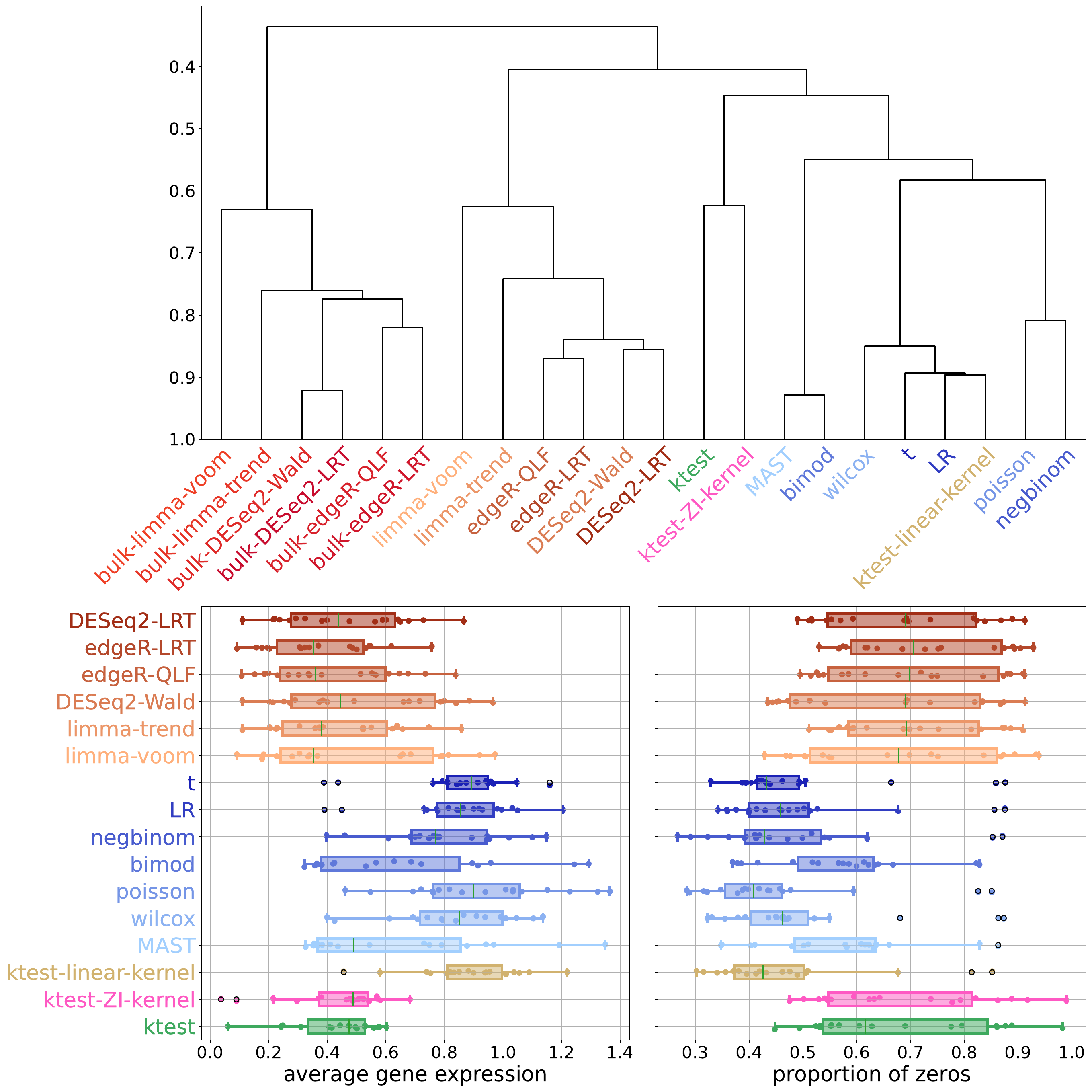} 
\end{center}
\caption{Top: Hierarchical clustering based on  average AUCC scores computed between pairs of methods (over 18 datasets \cite{squair_confronting_2021}). Bottom: Boxplot of the average expression (left) and proportion of zeros (right) of the top 500 DE genes for different DE methods (over 18 datasets \cite{squair_confronting_2021}). Red: bulk methods, orange: pseudobulk methods, blue: single-cell methods. \revised{The truncation parameter is set to $T=4$ for \texttt{ktest} (only univariate tests were performed).}}
\label{fig:squair}
\end{figure}

Differential analysis methods require validation through experimental data, typically by using a ground truth list of differentially expressed (DE) genes and an accuracy criterion. In this study, we examine the framework proposed by Squair et al. \cite{squair_confronting_2021}, which compared 14 DE analysis methods \revised{(Table \ref{tab:methods_squair})} on 18 scRNA-Seq datasets . The authors proposed three main conclusions: $i)$ replicate variability needs to be corrected, $ii)$ single-cell DE methods are susceptible to false discoveries, and $iii)$ pseudo-bulk methods are the most powerful. Pseudo-bulk methods involve applying DEA methods dedicated to bulk RNA-Seq to averaged scRNA-Seq. However, these conclusions are based on the use of bulk RNA sequencing DE genes as the ground truth, which inevitably favors pseudo-bulk methods designed to detect significant mean differences only. Hence, the study ignores genes with differential expression based on other characteristics, as shown in Korthauer's DB scenario \cite{korthauer_statistical_2016}. Therefore, we propose to broaden the scope of this comparative study by comparing the outputs of different DE methods in a pairwise comparative manner, without relying on a reference ground truth list of DE genes. Based on pair-wise accuracies, Differential Analysis methods cluster into three groups of concordant groups that correspond to bulk, pseudo-bulk and single-cell based methods respectively (Fig. \ref{fig:squair}, top). As expected, bulk-based methods are separated from others, pseudo-bulk and single-cell methods are clustered together because they are trained on scRNA-Seq data. Kernel testing shows performance close to single-cell methods. \revised{Kernel testing emerges as a third approach, aligning more closely with single-cell methods. Its top differentially expressed (DE) genes exhibit characteristics akin to those of pseudo-bulk methods in terms of average expression and the proportion of zeros. Notably, kernel testing diverges from other single-cell DEA methods, which typically identify highly-expressed genes, as highlighted in the original study (Fig. \ref{fig:squair}, bottom). It is noteworthy that when the kernel method employs a linear kernel, its performances are close to those of the $t$-test and likelihood-ratio test, illustrating the interest of a non-linear procedure.} By inspecting the distributional changes associated to genes considered as false-positive in the original study (with  bulk-RNA-Seq genes as the ground truth), we show that they can in fact be interpreted as true positives. Many of them belong to the DB alternative (difference in both modalities and proportions, \cite{korthauer_statistical_2016}), and were thus undetectable from bulk-RNA-Seq data and pseudo-bulk methods (Fig. \ref{fig:gene_distrib}, left). Their classification in terms of false positives is then questionable, and kernel testing is clearly powerful to detect those alternatives on experimental data. Others present slight shifts in distribution and low zero proportions, these genes are correctly detected by the ZI-Gauss kernel (examples of such distribution shapes are shown in Fig. \ref{fig:gene_distrib}, right). \revised{Finally we compared the computational time of competing methods, illustrating the quadratic complexity of \texttt{ktest} (Fig \ref{fig:gene_distrib}), which still remains reasonable for complete transcriptomes.}

\subsection{Kernel testing reveals the heterogeneity of reverting cells}

Single-cell transcriptomics has been widely used to investigate the molecular bases of cell differentiation, and has highlighted the stochasticity and dynamics of the underlying gene regulatory networks. The stochasticity of GRNs allows plasticity between cell states, and is a source of heterogeneity between cells along the differentiation path, which calls for multivariate differential analysis methods. We focus on the differentiation path of 
chicken primary erythroid progenitor cells (T2EC). A first study highlighted
the existence of plasticity, i.e. the ability of cells induced into differentiation to reacquire the phenotypic characteristics of undifferentiated cells (e.g. starting self-renewing again), until a differentiation point of commitment (around 24H after differentiation induction) after which this phenotype was lost \cite{richard_single-cell-based_2016}. A second study investigated the molecular mechanisms underlying cell differentiation and reversion by measuring cell transcriptomes at four time points : undifferentiated T2EC maintained in a self-renewal medium (0H), then put in a differentiation-inducing medium for 24h (24H). The population was then split into a first population maintained in the same medium for 24h to achieve differentiation (48HDIFF), the second population was put back in the self-renewal medium to investigate potential reversion (48HREV) \cite{zreika_evidence_2022}. Cell transcriptomes were measured using scRT-qPCR on 83 genes selected to be involved in the differentiation process, as well as scRNA-Seq to complement the study by a non-targeted approach. Despite the strong global transcriptomic similarity between 0H and 48HREV cells, four DE genes were identified in the study (\textit{RSFR}, \textit{HBBA}, \textit{TBC1D7}, \textit{HSP90AA1}), interpreted as either a delay or as traces of engagement into differentiation of the 48HREV population, before returning to the self-renewal state. Hence, these first analyses suggested some heterogeneities between undifferentiated cells and reverted cells. 

\begin{figure}
\begin{center}
\includegraphics[scale=.3]{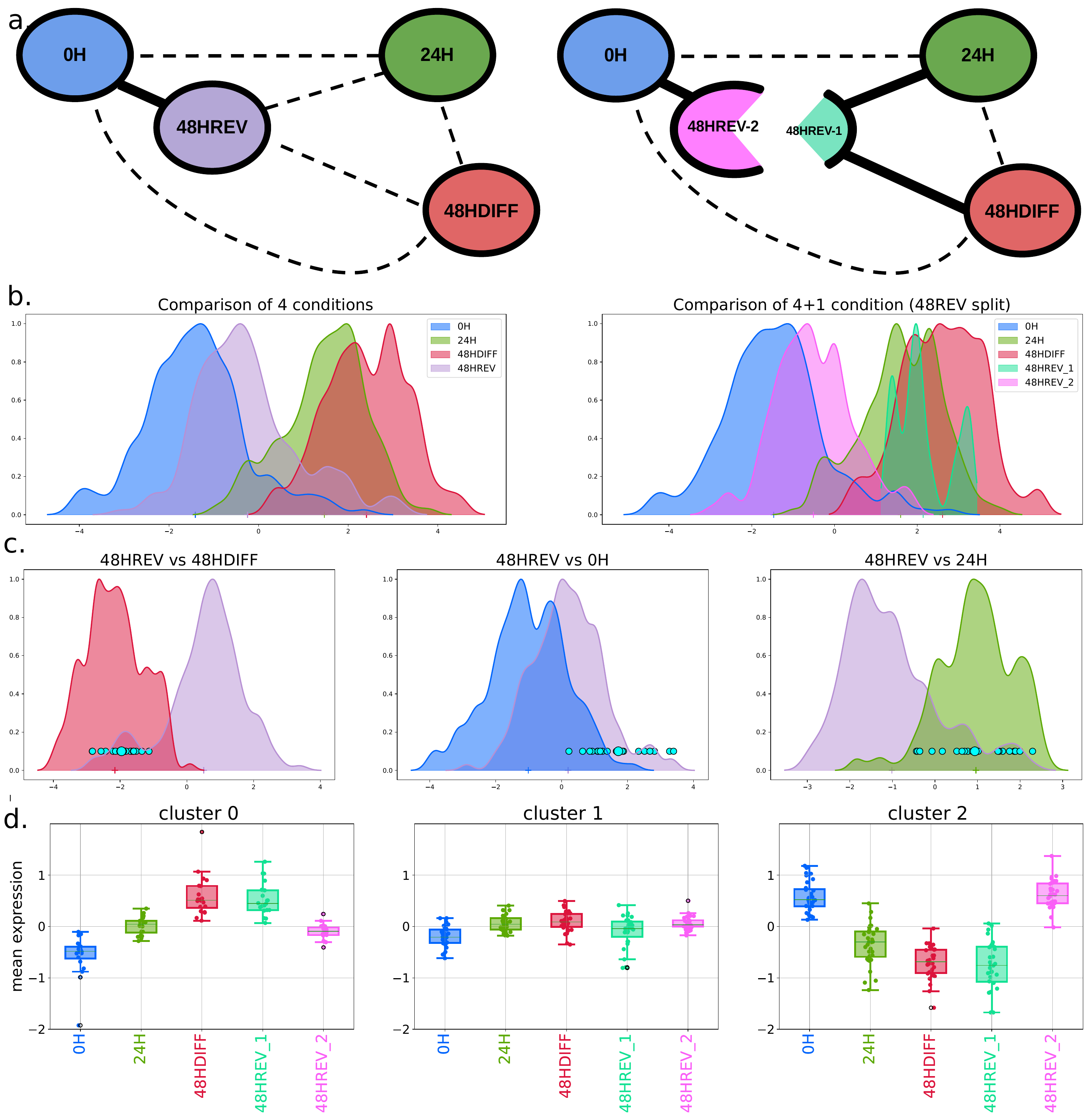}
\end{center}
\caption{a: Summarized distance graphs between conditions before (left) and after (right) splitting condition 48HREV into populations 48HREV-1 and 48HREV-2.
\revised{b: Cell densities of all compared conditions, before (left) and after (right) splitting condition 48HREV} c: Cell densities of compared conditions projected on the discriminant axis between conditions 48HREV and 48HDIFF (left), 48HREV and 0H (middle) and 48HREV and 24H (right)  with highlighted population 48HREV-1. d : \revised{Boxplots of the variation of the gene expression along the five populations 0H, 24H, 48HDIFF, 48HREV-1 and 48HREV-2 for the three genes clusters.} a,b,c and d are obtained from scRT-qPCR data. \revised{The multivariate differential expression analysis was performed with $T=10$.}}
\label{fig:reversion RTqPCR}
\end{figure}

\revised{Since the experiments were conducted on eight independent batches, our analysis began by assessing the significance of the batch effect using the multigroup kernel-based test. Both scRT-qPCR and scRNA-Seq data exhibited a significant effect (p-values of $3.18 \times 10^{-78}$ and $1.26 \times 10^{-85}$, respectively). To address this, we corrected the data embedding by applying the mean embedding of the batch effect, resulting in a non-linear normalization with respect to the batch (details in the Supplementary Material). Then we conducted a new test to compare the batch-corrected distribution of gene expressions between biological conditions (differentiation time). The multigroup kernel test first confirmed heterogeneity among conditions in both scRT-qPCR and scRNA-Seq (p-values of $0$ and $3.64 \times 10^{-142}$, respectively). The 4-group discriminant analysis yielded three discriminant axes that represent the global heterogeneities of the data. Notably, the first discriminant axis associated with the global 4-group comparison ordered the four conditions according to the time of differentiation (Fig. \ref{fig:reversion RTqPCR}.b and \ref{fig:reversion scRNAseq}), while subsequent axes provided less pronounced information (Fig. \ref{fig:reversion scRNAseq multigroupe}). We then employed \texttt{ktest} for pair-wise comparisons between conditions, confirming a significant difference between undifferentiated cells (0H) and reverted cells (48HREV) in both scRT-qPCR and scRNA-Seq data (p-values of $4.55 \times 10^{-23}$ and $7.39 \times 10^{-06}$, respectively). However, considering the test statistic as a distance also confirmed the close proximity between these two conditions (Fig. \ref{fig:reversion RTqPCR multigroupe} and \ref{fig:reversion scRNAseq multigroupe}).}


We assumed that population 48HREV was heterogeneous and contained reverted cells and non-reverted cells. A $k$-means clustering was unable to detect any particular cell cluster (Fig. \ref{fig:reversion justification}, middle). As the discriminant axis provided by our framework represents a synthetic summary of the global transcriptomic differences between two cell populations, it allowed to highlight the existence of a sub-population of 48HREV cells (denoted 48HREV-1) that overlaps  the distribution summary of 48HDIFF-cells (48HREV vs. 48HDIFF, Fig. \ref{fig:reversion RTqPCR}.c). Interestingly, these cells also matched the distribution summary of 24H-cells (48HREV vs. 24H, Fig. \ref{fig:reversion RTqPCR}.c), and were separated from the undifferentiated cells (48HREV vs 0H, Fig. \ref{fig:reversion RTqPCR}.c). A similar sub-population was detected using scRNA-Seq data (48HREV vs. 48HDIFF Fig. \ref{fig:reversion scRNAseq}.b). According to our test, populations 48HDIFF and 48HREV-1 were very slightly different on scRT-qPCR data and similar on scRNA-Seq data ($p$-values $4.73$ $10^{-5}$ and $0.80$ respectively). This slight difference may be explained by the targeted approach of scRT-qPCR that was based on a selection of 83 genes involved in differentiation and on the higher precision of the scRT-qPCR technology \cite{zreika_evidence_2022}. 48HREV-2 cells (48HREV cells after removing 48HREV-1 cells) were closer but still significantly different from 0H cells in both technologies ($p$-values $4.48$ $10^{-17}$ and $3.98$ $10^{-05}$ respectively).
To describe these two sub-populations in terms of genes, we performed a $k$-means clustering on the averaged centered expressions of genes over cells in populations 0H, 24H, 48HDIFF, 48HREV-1, 48HREV-2. We identified three and five gene clusters on the scRT-qPCR and the scRNA-Seq data respectively. These clusters can be separated in three groups (Fig \ref{fig:reversion RTqPCR}.d and \ref{fig:reversion scRNAseq}.c): 

(\textit{i}) genes activated during differentiation (scRT-qPCR cluster 0, scRNA-Seq clusters 2 and 3), e.g. hemoglobin related genes such as \textit{HBA1} and \textit{HBAD} (shown in Fig. \ref{fig:reversion scRNAseq}.d), (\textit{ii}) genes deactivated during differentiation (scRT-qPCR cluster 2, scRNA-Seq cluster 0) e.g. genes involved in metabolism of self-renewing cells such as \textit{LDHA} and \textit{LY6E} (shown in Fig. \ref{fig:reversion scRNAseq}.d), and (\textit{iii}) genes with no clear function pattern for which the expression levels did not change much during differentiation and reversion (scRT-qPCR cluster 1 and scRNA-Seq clusters 1 and 4). The $p$-value tables associated to each pair-wise univariate DE analysis with respect to each gene cluster are available online \footnote{\url{https://github.com/AnthoOzier/kernel_testsDA.git}}.   

To conclude, our differential transcriptome framework showed that population 48HREV is composed of two sub-populations, which sheds light on new putative mechanisms driving differentiation and reversion processes. Whereas a population is only slightly different to undifferentiated cells (48HREV-2), a sub-population (48HREV-1) has remained engaged in differentiation. This difference could be either due to a delay in engaging the reversion process for some cells, or to cells having crossed the irreversible point of commitment. Furthermore, our method has identified cellular pathways which could be important either for cell plasticity or cell differentiation, and can guide design of further experiments. Overall, it could enhance our comprehension of how gene regulatory networks react to differentiation and reversion signals.

\section{Towards a new testing framework for differential binding analysis in single-cell ChIP-Seq data}

There is currently no dedicated method for the comparison of single-cell epigenomic profiles, existing studies often use non-parametric testing to compare epigenomic states and retrieve differentially enriched loci. The joint multivariate testing strategy seems particularly suited to compare epigenomic data since it is well established that chromatin conformation and natural spreading of histone modifications - in particular H3K27me3 \cite{margueron_role_2009} - can induce complex dependencies between sites occupancy. A recent study \cite{marsolier_h3k27me3_2022} has shown that the repressive histone mark H3K27me3 (trimethylation of histone H3 at lysine 27) is involved in the emergence of drug persistence in breast cancer cells. Drug persistence occurs when only a subset of cells, known as persister cells, survives the initial drug treatment, thereby creating a reservoir of cells from which resistant cells will emerge. The study identified a persister expression program involving genes such as \textit{TGFB1} and \textit{FOXQ1}, with H3K27me3 as a lock to its activation. Changes in H3K27me3 modifications at the single-cell level showed a consistent pattern in persister cells compared to untreated cells, in particular cells display recurrent losses of repressive histone methylation at a subset of genes of the persister expression program. However, this pattern was not necessarily maintained in cells that developed full resistance, suggesting the that part of the epigenomic features of persister cells might be transient. Moreover, analysis of untreated cells revealed heterogeneity within epigenomic profiles. Part of the population exhibited shared epigenomic features with persister cells, yet remaining distinguishable from them. This initial analysis suggested that a pool of untreated cells could contribute to the persister cell population later upon exposure to chemotherapy. However, unsupervised analyses were unable to clearly identify this pool of precursor cells.

\begin{figure}
\begin{center}
\includegraphics[scale=0.9]{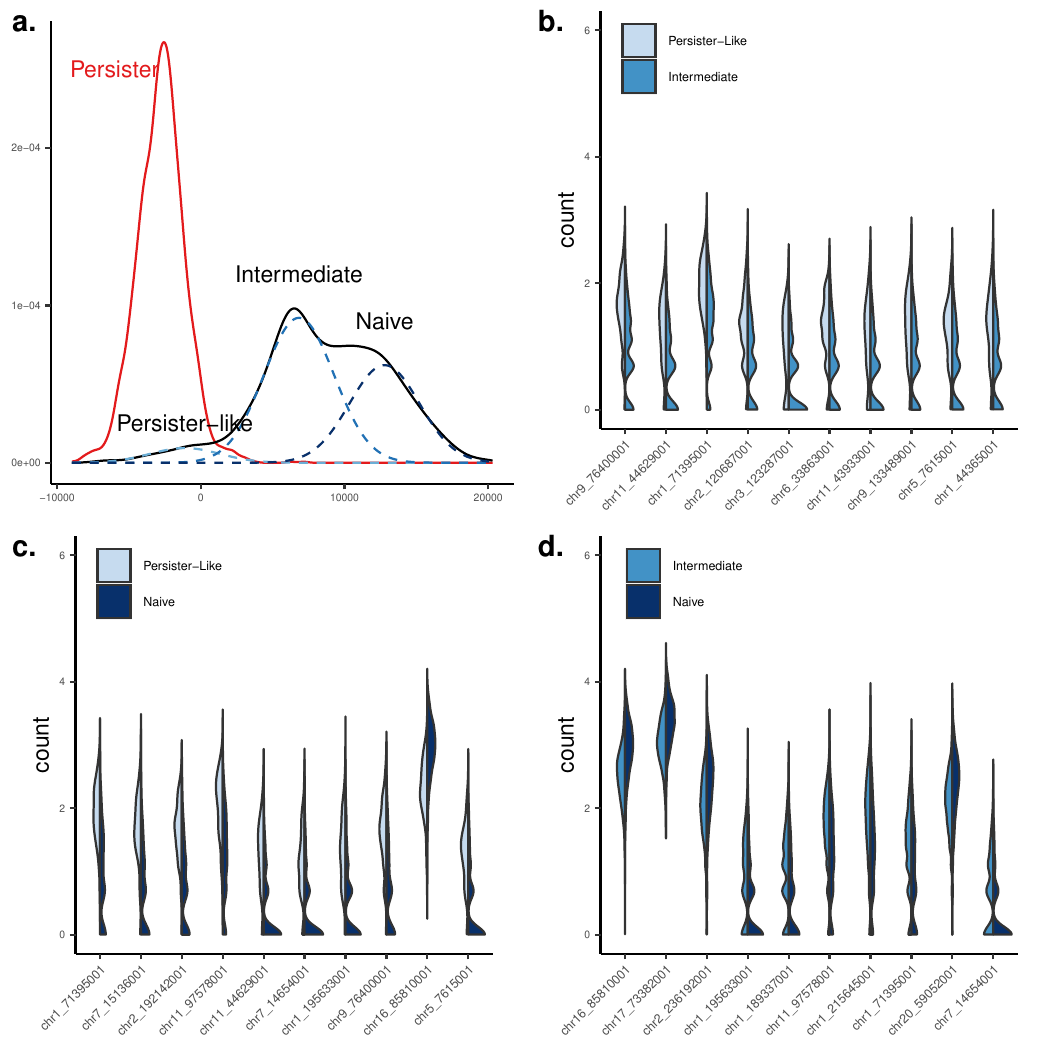}
\end{center}
   \caption{Differential analysis of scChIP-Seq data on breast cancer cells. a. Cell densities of persister cells vs. untreated cells. Sub-populations of untreated cells were identified using 3-component mixture model, that revealed persister-like cells, intermediate and naive cells. b-c-d : violin plots of the top-10 differentially enriched H3K27me3 loci between the 3 sub-populations. Features are designated by the genomic coordinates of the ChIP-Seq peaks. Corresponding overlapping genes are provided in Table \ref{tab:chip}. \revised{Multivariate (a) and univariate analyses (b-c-d) were performed with $T=5$.}}
\label{fig:chipseq}
\end{figure}

We compared H3K27me3 scChIP-Seq profiles between untreated and persister cells using kernel testing. Thanks to the discriminative approach, our framework offers a synthetic representation of the distributional differences between cell populations \ref{fig:chipseq}. Projecting cells on the kernelized discriminant axis reveals 3 sub-populations within the untreated cell population: Persister-Like (109 cells; 5\% of untreated cells), Intermediate (1124 cells; 57\%), Naive (744 cells; 38\%), with increasing distance to persister cells (Fig. \ref{fig:chipseq}). We then performed a differential analysis of H3K27me3 enrichment between persister cells and the $n=109$ untreated cells that were the most similar to persister cells on the discriminant axis. Over the 6,376 tested regions, only 14 were significantly differentially enriched ($p$-value$<10^{-3}$, Table \ref{tab:chip}), suggesting that this sup-population of untreated cells is epigenomically very close to persister cells (with persister cells being hypo-methylated on these significant regions compared to persister-like cells). We then studied the differences between the three populations present in the untreated cell population, prior to any treatment. We performed differential analysis between the most distant untreated cells ('naive' vs 'intermediate'), and between 'intermediate' cells and 'persister-like' cells. We detected significant changes in repressive epigenomic enrichments, both losses and gains, that will need further functional testing to understand their potential role in drug-persistence. Altogether, our new kernel analytical framework shows that persister-like cells could exist prior to any treatment, and provides a novel level of appreciation of epigenomic heterogeneity - by revealing three sub-populations within treatment naive cell population. In addition, our method identifies small quantitative variations that are not detected by other methods and will need to be related to gene expression and other molecular features for further interpretation.

\section{Conclusion}

In this work we introduced the framework of kernel testing to perform differential analysis in a non-linear setting. This method compares the distribution of gene expression or epigenomic profiles through global or feature-wise comparisons, but can be extended to any measured single-cell features. Kernel testing has focused much attention in the machine learning community since it has the advantage of being non-linear, computationally tractable, and provides visualization combining dimension reduction and statistical testing. Its application to single-cell data is particularly promising, as it allows distributional comparisons without any assumptions about their shape. Moreover, using a classifier to perform discrimination-based testing has become popular \cite{kim_classification_2021}, and allows powerful detection of population heterogeneities in both expression and epigenomics single-cell data. Our simulations show the power of this approach on specifically designed alternatives \cite{korthauer_statistical_2016}. Furthermore, comparing kernel testing with other methods based on multiple scRNA-Seq data reveals its superior capability to identify distributional changes that go undetected by other approaches. Finally, the application of kernel testing to scRNA-Seq and scChIP-Seq data uncovers biologically meaningful heterogeneities in cell populations that were not identified by standard procedures. \revised{We also demonstrate the applicability of kernel-testing for multiple group comparisons and two-factor designs. Our plan is to fully develop this approach, providing a comprehensive mathematical framework that facilitates the study of any complex design, including model validation and contrast testing, for instance.} More than ever, single-cell data science appears at the convergence of many cutting-edge methodological developments in machine learning. As a result, these advancements will have significant implications for the old-tale of differential analysis, offering new avenues for progress and improvement.



\section{Materials and Methods}

\subsection{Simulation settings}

The comparison study on data simulated was performed on data following different mixtures of zero inflated negative binomial (ZINB) distributions \cite{gauthier_distribution-free_2021}. The distribution parameters were chosen to reproduce the four Korthauer alternatives and two types of $H_0$ distributions. The performances were computed on 500 repetitions of a dataset composed of 1000 DE genes and 9000 non-DE genes. The DE genes are equaly separated in the four alternatives DE,DM, DP and DB. The non-DE genes are equally separated into an unimodal ZINB and a bimodal mixture of ZINB. The DE methods were applied on the raw data, type-I errors and powers were computed on the raw p-values while  false discovery and true discovery rates were computed on the adjusted $p$-values, with the Benjamini-Hochberg correction \cite{benjamini_et_hochberg_controlling_1995}. \revised{Compared methods are shown in Table \ref{tab:methods_simu}.}

\subsection{Comparison of methods on published scRNA-Seq}
The eighteen comparison datasets were downloaded from the Zenodo repository (\url{https://doi.org/10.5281/zenodo.5048449}) compiled by Squair and coauthors \cite{squair_confronting_2021}. They consists of six comparisons of bone marrow mononuclear phagocytes from mouse, rat, pig and rabbit in different conditions \cite{hagai_gene_2018}, eight comparisons of naive and memory T cells in different conditions \cite{cano-gamez_single-cell_2020} and four comparisons of alveolar macrophages and type II pneumocytes between young and old mouses \cite{angelidis_atlas_2019} and between patients with pulmonary fibrosis and control individuals \cite{reyfman_single-cell_2019}. More details on the datasets are in \cite{squair_confronting_2021} or in the original studies. The preprocessing step consisted in filtering genes present in less than three cells and normalizing data with the Seurat function \textit{NormalizeData}, as in the original comparative study \cite{squair_confronting_2021}. This not very restrictive preprocessing was chosen in order to not introduce biaises in the analyses, and many genes would have been ignored form the analysis in real conditions. The Area Under the Concordance Curves (AUCC) scores were computed with the original scripts \cite{squair_confronting_2021}. 

\subsubsection{Zero-Inflated Gaussian Kernel}

\revised{Our method is non-parametric, meaning we do not assume a specific distribution for the data. In this context, we propose to derive a kernel that is tailored to a high proportion of zeros. To achieve this, we propose to develop the zero-inflated Gauss kernel, which involves considering a zero-inflated Gaussian distribution with $\pi$ the proportion of additional zeros:
\begin{eqnarray*}
X \sim \pi \delta_0(\bullet) + (1-\pi) f_{\mu,\sigma}(\bullet),
\end{eqnarray*}
with $f_{ \mu,\sigma}$ the Gaussian probability density function. It is important to note that this does not imply that we assume the data to follow a zero-inflated Gaussian distribution. This representation serves merely as a methodological framework for deriving the new kernel.} This distribution has a mixture representation, with $Z$ standing for the binary variable of distribution $\mathcal{B}(\pi)$, such that
\begin{eqnarray*}
f_{\mu,\sigma,\pi}(x) = \mathbb{P}(Z=1)  \delta_0(x) +  \mathbb{P}(Z=0) f_{\mu,\sigma}(x)
\end{eqnarray*}
We know the probability kernels for the Gaussian part of the model:
\begin{eqnarray*}
k_{\text{Gauss}}(\mu,\mu') = \frac{1}{4\pi \sigma^2} \exp ( -(\mu- \mu')^{2} / 4\sigma^2)
\end{eqnarray*}
and for the Bernoulli distribution:
\begin{eqnarray*}
k_{\mathcal{B}}\Big( \pi,\pi' \Big)= \pi \pi' + (1-\pi)(1-\pi').
\end{eqnarray*}
To get the ZI-Gauss kernel, we compute the probability densities $f_{\mu,\sigma,\pi}$ and $f_{\mu',\sigma,\pi'}$
\begin{eqnarray*}
k_{\text{ZI-Gauss}} \Big(f_{\mu,\sigma,\pi},f_{\mu',\sigma,\pi'} \Big) 
&=& \int_{x} \left( \sum_{z}  \Big( \mathbb{P}_{\pi}(Z=z) f_{\mu,\sigma,\pi}(x \mid Z=z)\Big) \right) \left( \sum_{z} \Big( \mathbb{P}_{\pi'}(Z=z) f_{\mu',\sigma,\pi'}(x \mid Z=z)\Big) \right) dx \\
& =& \pi \pi' + \pi(1-\pi')f_{ \mu',\sigma}(0) + (1-\pi)\pi'f_{ \mu,\sigma}(0) +(1-\pi)(1-\pi') k_{\text{Gauss}}(\mu,\mu'),
\end{eqnarray*}
 In the simulations, the ZI-Gauss kernel was computed using the parameters of the Binomial distributions used to determine the drop-out rates of the simulated data (drawn uniformly in $[0.7, 0.9]$), the variance parameter $\sigma$ was set as the median distance between the non-zero observations and the Gaussian means $\mu$ were set as the observed values.

\subsection{Reversion data}
Details on the experiment and on the data can be found in the original paper \cite{zreika_evidence_2022}. The kernel-based testing framework was performed on the $\log(x+1)$ normalized RT-qPCR data and on the Pearson residuals of the 2000 most variable genes of the scRNA-Seq data obtained through the R package \texttt{sctransform} \cite{hafemeister_normalization_2019}. \revised{For both datasets, we corrected for the batch effect in the feature space.} \revised{The gene clusters were computed on the data after correcting for the batch effect in the input space.} The truncation parameter of the global comparisons ($T=10$ for both technologies) was chosen to be large enough for the discriminant analysis to capture enough of the multivariate information and to maximize the discriminant ratio. The truncation parameter retained for univariate testing ($T=4$) was chosen according to the simulation study.

\subsection{sc-chIP-Seq data}

Single-cell chIP-Seq data correspond to a count matrix of unique reads mapping to the genome binned into H3K27me3 previously identified peaks \cite{marsolier_h3k27me3_2022}. This matrix was filtered for cells with a minimum coverage of 3,000 unique reads and a maximum coverage of 10,000 reads. Top $5\%$ covered cells were furthered filtered out, as potential doublets

\section{Declarations}
\subsection{Ethics approval and consent to participate}
Not applicable
\subsection{Consent for publication}
Not applicable

\subsection{Availability of data and materials}

The data used to compare methods are available from the Zenodo repository (\url{https://doi.org/10.5281/zenodo.5048449}) as compiled by Squair and coauthors \cite{squair_confronting_2021}. Reversion scRT-qPCR data are available in the SRA repository number SRP076011, and fully described in the original publication \cite{zreika_evidence_2022}. Single-cell chIP-Seq data can be found on GEO with the accession number GSE164385 \cite{marsolier_h3k27me3_2022}.

\subsection{Competing interests}

The authors declare that they have no competing interests

\subsection{Funding}

The research was supported by a grant from the Agence Nationale de la Recherche ANR-18-CE45-0023 SingleStatOmics, by the projects AI4scMed, France 2030 ANR-22-PESN-0002, and SIRIC ILIAD (INCA-DGOS-INSERM-12558). 

\subsection{Authors' contributions}

AOL, BM, FP developed the method, analyzed the data and wrote the manuscript, AOL, GD, PA developed the python/R \texttt{ktest} package, CV participated to the analysis of epigenomics data, CF, OG and SGG participated to the analysis of the scRNA-Seq reversion data. BM and FP supervised the project.

\subsection{Acknowledgements}

The authors would like to thank Boris Hejblum for sharing the simulated data, François Gindraud for helping on the implementation of the kernel method, Stéphane Minvielle and  Zaid Harchaoui for fruitful scientific discussions. This work was performed using HPC resources from GLiCID computing center.

\newpage

\appendix

\renewcommand{\thesection}{S.\arabic{section}}
\renewcommand{\thesubsection}{S.\arabic{subsection}}
\renewcommand{\thetable}{S.\arabic{table}}
\renewcommand{\thefigure}{S.\arabic{figure}}
\renewcommand{\theequation}{S.\arabic{equation}}

\section{Supplementary Material}

\subsection{Full derivation of the Maximum Mean Discrepancy}
\revised{We provide the full derivation of the Maximum Mean Discrepancy to explain how expected intra and inter-condition distances are involved in the definition of the statistic.
\begin{eqnarray*}
\operatorname{MMD}^2(\mu_1,\mu_2) &=& \|\mu_1 - \mu_2\|^2_{\mathcal{H}} \\
&=& \langle \mu_1, \mu_1\rangle +  \langle \mu_2, \mu_2\rangle - 2  \langle \mu_1, \mu_2\rangle \\
&=& \langle \mathbb{E}_{Y_1 \sim \mathbb{P}_1}\big( \phi(Y_1) \big) , \mathbb{E}_{Y'_1 \sim \mathbb{P}_1}\big( \phi(Y'_1) \big)\rangle +  \langle \mathbb{E}_{Y_2 \sim \mathbb{P}_2}\big( \phi(Y_2) \big), \mathbb{E}_{Y'_2 \sim \mathbb{P}_2}\big( \phi(Y'_2) \big)\rangle \\
&-& 2  \langle \mathbb{E}_{Y_1 \sim \mathbb{P}_1}\big( \phi(Y_1) \big), \mathbb{E}_{Y_2 \sim \mathbb{P}_2}\big( \phi(Y_2) \big)\rangle \\
&=&  \mathbb{E}_{Y_1 \sim \mathbb{P}_1}  \mathbb{E}_{Y'_1 \sim \mathbb{P}_1} \langle \phi(Y_1)  , \phi(Y'_1) \rangle +  
     \mathbb{E}_{Y_2 \sim \mathbb{P}_2}  \mathbb{E}_{Y'_2 \sim \mathbb{P}_2} \langle \phi(Y_2)  , \phi(Y'_2) \rangle \\
&-& 2  \mathbb{E}_{Y_1 \sim \mathbb{P}_1}  \mathbb{E}_{Y_2 \sim \mathbb{P}_2} \langle \phi(Y_1)  , \phi(Y_2) \rangle  \\
&=&\mathbb{E}_{Y_1 \sim \mathbb{P}_1, Y_1' \sim \mathbb{P}_1} \left[ k(Y_1,Y_1') \right] + \mathbb{E}_{Y_2 \sim \mathbb{P}_2, Y_2' \sim \mathbb{P}_2} \left[ k(Y_2,Y_2') \right] \\
&-& 2 \times \mathbb{E}_{Y_1 \sim \mathbb{P}_1, Y_2 \sim \mathbb{P}_2} \left[ k(Y_1,Y_2) \right].
\end{eqnarray*}}

\subsection{Generalization to the multiple-conditions comparisons}
\revised{Our approach can be generalized to the comparison of $I$ conditions. Consider $I\geq2$ groups of $n_1,\dots,n_I$ observations (with $\sum_{i=1}^{I}n_i=n$) sucht that :
\begin{eqnarray*}
Y_{i,j} \sim \mathbb{P}_i, \quad i = 1,\dots,I \quad j = 1, \dots n_i.
\end{eqnarray*}
For $i\in\{1,\dots,I\}$, we denote by $\mu_i$ the kernel mean embedding of distribution $\mathbb{P}_i$ and $\mu = \sum_{i=1}^{I} n_i/n \mu_i$ is the kernel mean embedding of the distribution associated with the complete data. We can define the within-group covariance operator $\Sigma_W$ and the between-group covariance operator $\Sigma_B$ such that: 
\begin{eqnarray*}
\Sigma_{W} &=& \sum_{i=1}^I \frac{n_i}{n} \Sigma_i \\ 
\Sigma_{B} &=& \sum_{i=1}^I \frac{n_i}{n^2} (\mu_i - \mu)^{\otimes 2}.
\end{eqnarray*}
Then, the test statistic has the same expression as the two-sample test statistic and asymptotically follows a Chi-square distribution with $(I-1) \times T$ degrees of freedom (\cite{thesis_Anthony}, Chap 3). This discriminant approach with $I$ conditions has $(I-1)$ discriminant directions defined as the $(I-1)$ first eigen-directions of the operator $(\Sigma_{W,T}^{-1} \Sigma_B)$. Note that when $I>2$, the discriminant directions cannot be written explicitly with respect to the kernel mean embeddings, as in the two-sample case, which implies that the test statistic cannot be rewritten as a Mahalanobis distance.}

\subsection{Tuning the truncation hyperparameter}

\begin{figure}
\begin{center}
\includegraphics[scale=0.25]{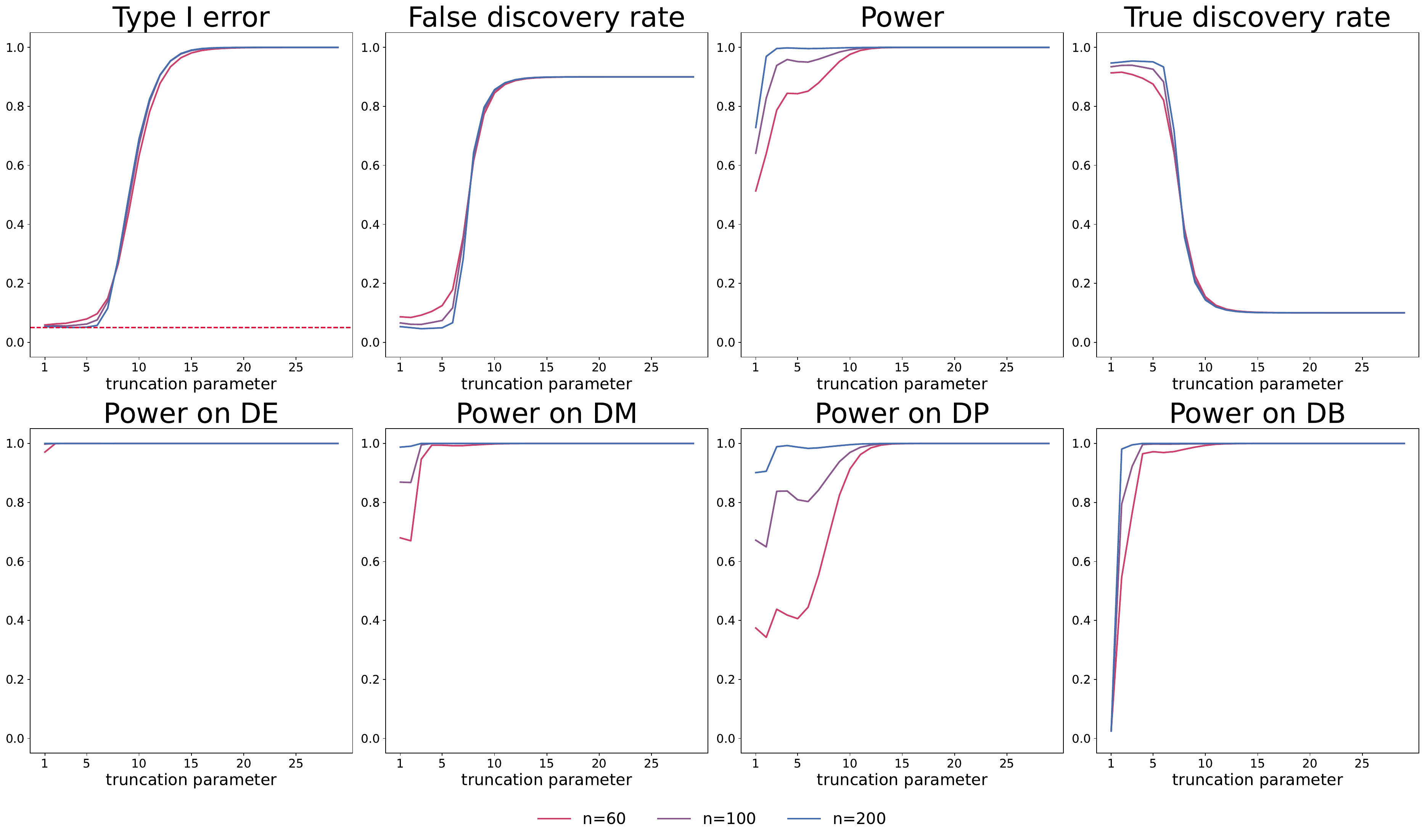}
\end{center}
   \caption{Calibration of the truncation with respect to type-I errors and power. Top: Type-I errors are computed on raw $p$-values under $H_0$. False discovery Rate computed on Benjamini-Hochberg adjusted $p$-values. Power computed on raw $p$-values under $H_1$. True Discovery Rate computed on Benjamini-Hochberg adjusted $p$-values. Simulated data consists of $10000$ genes ($1000$ DE, $9000$ non-DE). Alternatives are simulated using DE : classical difference in expression ($250$ genes), DM : difference in modalities ($250$ genes), DP : difference in proportions ($250$ genes), DB : difference in both modalities and proportions with equal means ($250$ genes). Error rates are computed over $500$ replicates.}
\label{fig:performances_ccdf_t}
\end{figure}

We use the simulation data to calibrate the hyperparameter of our method, i.e. the number $T$ of principal directions of the within-covariance operator to retain to  regularize the kernel-based Mahalanobis distance. The theoretical calibration of this hyperparameter still requires heavy mathematical developments, as shown by recent work  \cite{hagrass_spectral_2022}. However, these simulations provide a simple rule of thumb to choose it. Indeed, since $T$ can be interpreted as the quantity of within-variance information used to describe the residual expression, increasing $T$ will increase power in the detection of complex alternatives, at the price of increased type-I errors. In the simulations, Type-I errors of the kernel test remains at the nominal level $\alpha=5\%$ until $T\leq 6$. with maximal power for $T=4$ (Fig \ref{fig:performances_ccdf_t}). Interestingly, the test was completely unable to detect the DB alternative when $T=1$. These results confirm that the truncation hyperparameter should be chosen as a trade-off between maximizing testing power while keeping the type-I errors controlled at the nominal level to ensure calibration. \revised{This motivates the choice of $T=4$ for the univariate DE analyses in the simulations and the sc-RNASeq application. $T=5$ was chosen for the sc-chIPSeq example}. 

For multivariate analyses, we assumed that the meaningful information was contained in more than four principal directions of the within-covariance operator and chose to take a larger truncation parameter in order to take into account more of the multivariate information available. We then chose the truncation parameter $T=10$ \revised{for the sc-RNASeq example and $T=5$ for the sc-chIPSeq data,} that maximized the discriminant ratio while being not too large to still ensure the calibration.

\subsection{Kernel trick for the effective computation of the test statistic}
In this section, we describe how to compute the test statistic $\widehat{D}_T^2(\widehat{\mu}_1,\widehat{\mu}_2)$ and the vector of projections of the embeddings onto the discriminant axis $V$, with $i \in \{1,2\}$, $j \in \{ 1, \dots, n_i \}$, and $V = (\left \langle h_T^\star , \phi(Y_{i,j}) \right \rangle_{\mathcal{H}})_{i,j}$ for $T \in \{ 1 , \dots, n\}$. This computation relies on the kernel trick that consists in expressing every quantity of interest with respect to the gram matrix $K$ containing every pair-wise evaluation of the kernel function, such that for $i,i^\prime \in \{1,2\}$, $K = (K_{i,i'})_{i,i'}$, where for  $j \in \{ 1, \dots, n_i \}$, $j^\prime \in \{ 1, \dots, n_{i^\prime} \}$, $K_{i,i^\prime} = (k(Y_{i,j},Y_{i^\prime,j^\prime}))_{j,j'}$. The computation has two steps. First, we determine a matrix $K_W$ that has the same eigenvalues as the operator $\widehat{\Sigma}_W$, then we compute the quantities of interest with respect to $K$, the $T$ first eigenvalues $(\widehat \lambda_t)$, $t \in \{ 1,\dots,T \}$ and the associated unit eigenvectors $(u_t)_{t}$ of $K_W$. Let's denote by $I_n$ the identity matrix of size $n$, $J_n$ the matrix of size $n$ full of $1$, and $\mathds{1}_n$ the vector of size $n$ full of $1$. Then for $i \in \{1,2\}$, let $P_i = I_{n_i} - n_i^{-1} J_{n_i}$, $P = \operatorname{diag}( P_1, P_2 )$ and $\omega = (n_1^{-1}\mathds{1}_{n_1}, -n_2^{-1} \mathds{1}_{n_2})^\prime \in \R^n$. We can show that the matrix $K_W$ is equal to $K_W = n^{-1} P K P$.
Then we have :
\begin{eqnarray*}
\widehat{D}_T^2(\widehat{\mu}_1,\widehat{\mu}_2) = \frac{n_1n_2}{n^2}   \sum_{t = 1}^{T} \widehat {\lambda}_t^{-2} (u_t^\prime P K \omega)^2, \quad \text{and} \quad 
V = \frac{n_1n_2}{n^2} \sum_{t = 1}^{T} \widehat{\lambda}_t^{-2} (u_t^\prime P K \omega ) K P u_t.
\end{eqnarray*}
\revised{To correct for a batch effect, consider the case where the design shows $B$ batches such that for $i\in\{1,2\}$ and $b\in\{1,\dots,B\}$, we have $n_{i,b}$ observations of condition $i$ in batch $b$ and $n_b = n_{1,b} + n_{2,b}$, then, for $i\{1,2\}$, we denote $Y_i = (Y_{i,1,1},\dots,Y_{i,1,n_{1,1}},\dots,Y_{i,B,1} ,\dots,Y_{i,B,n_{i,B}})$. To correct the batch effect in the feature space we considering the embedding $\widetilde{\phi}(Y_{i,b,j}) = \phi(Y_{i,b,j}) - n_{b}^{-1} \sum_{i^\prime=1}^{2}\sum_{j^\prime=1}^{n_{i^\prime,b}}\phi(Y_{i^\prime,b,j^\prime})$ for $i\in\{1,2\},b\in\{1,\dots,B\}$. In practice, this is done by replacing the Gram matrix $K$ by the matrix $P_B K P_B$, where $P_B = (Q_{(i,b),(i^\prime,b^\prime)})_{i,i^\prime\in\{1,2\},b,b^\prime\in\{1,\dots,B\}}$, with 
\begin{eqnarray*}
Q_{(i,b),(i^\prime,b^\prime)}
& =& \left\{
    \begin{array}{ll}
        I_{n_{i,b}} - n_b^{-1} J_{n_{i,b}}
        & \mbox{if } (i,b) = (i^\prime,b^\prime) \\
        - n_b^{-1}  J_{n_{i,b},n_{i',b}} & \mbox{if } i\neq i^\prime \mbox{and } b=b^\prime \\
       0 & \mbox{otherwise.}
    \end{array}
\right.
\end{eqnarray*}}

\begin{figure}
\begin{center}
\includegraphics[scale=0.25]{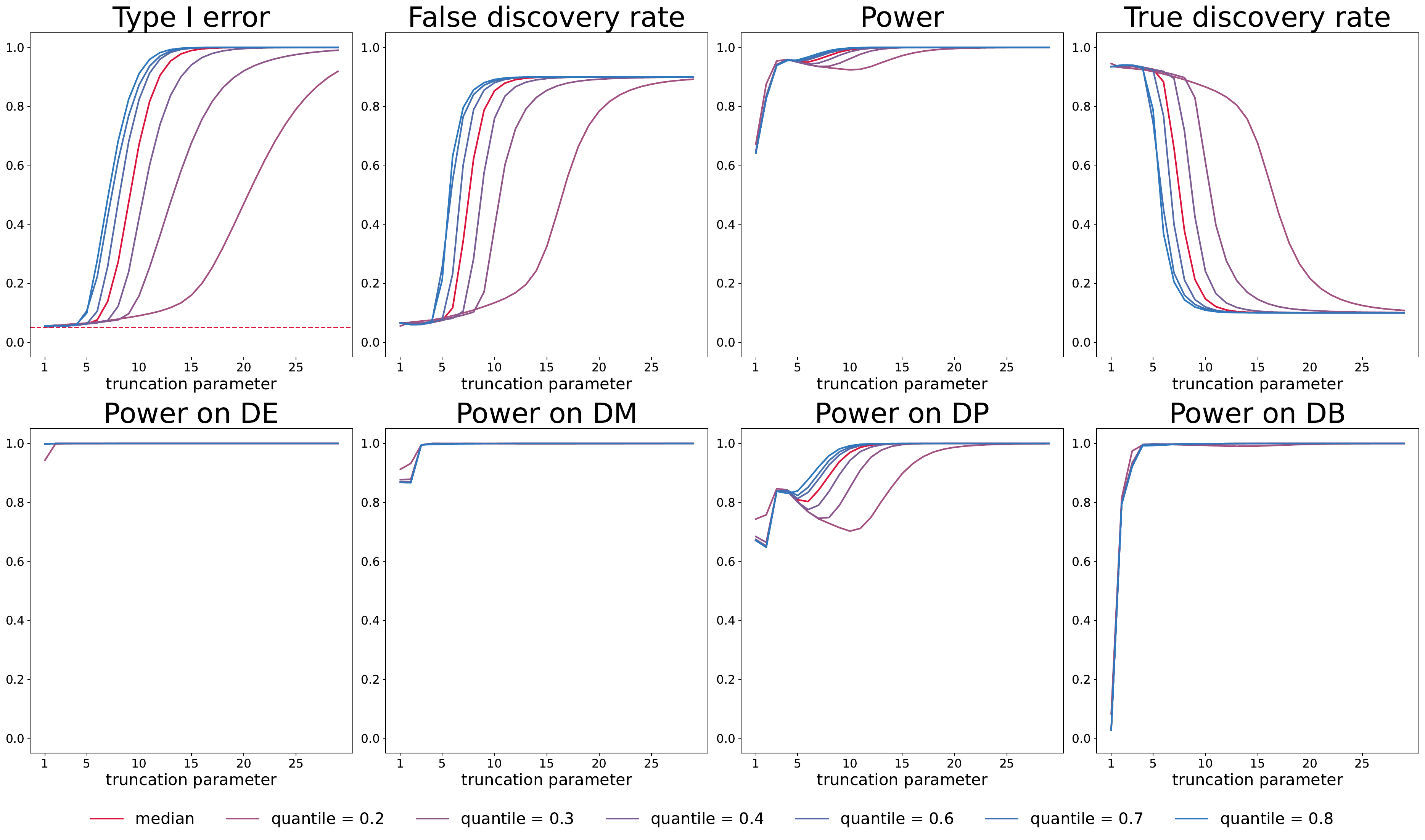}
\end{center}
   \caption{\revised{Impact of the kernel's bandwidth on error-rates and power.} Top: Type-I errors are computed on raw $p$-values under $H_0$. False discovery Rate computed on Benjamini-Hochberg adjusted $p$-values. Power computed on raw $p$-values under $H_1$. True Discovery Rate computed on Benjamini-Hochberg adjusted $p$-values. Simulated data consists of $10000$ genes ($1000$ DE, $9000$ non-DE). Alternatives are simulated using DE : classical difference in expression ($250$ genes), DM : difference in modalities ($250$ genes), DP : difference in proportions ($250$ genes), DB : difference in both modalities and proportions with equal means ($250$ genes). Error rates are computed over $500$ replicates. \revised{The bandwidth is computed as the quantiles of the pairwise distances (as explained in Section \ref{Section:kernel_choice} for the median).}
   \label{fig:bandwidth}
}
\end{figure}

\begin{figure}
\begin{center}
\includegraphics[scale=0.22]{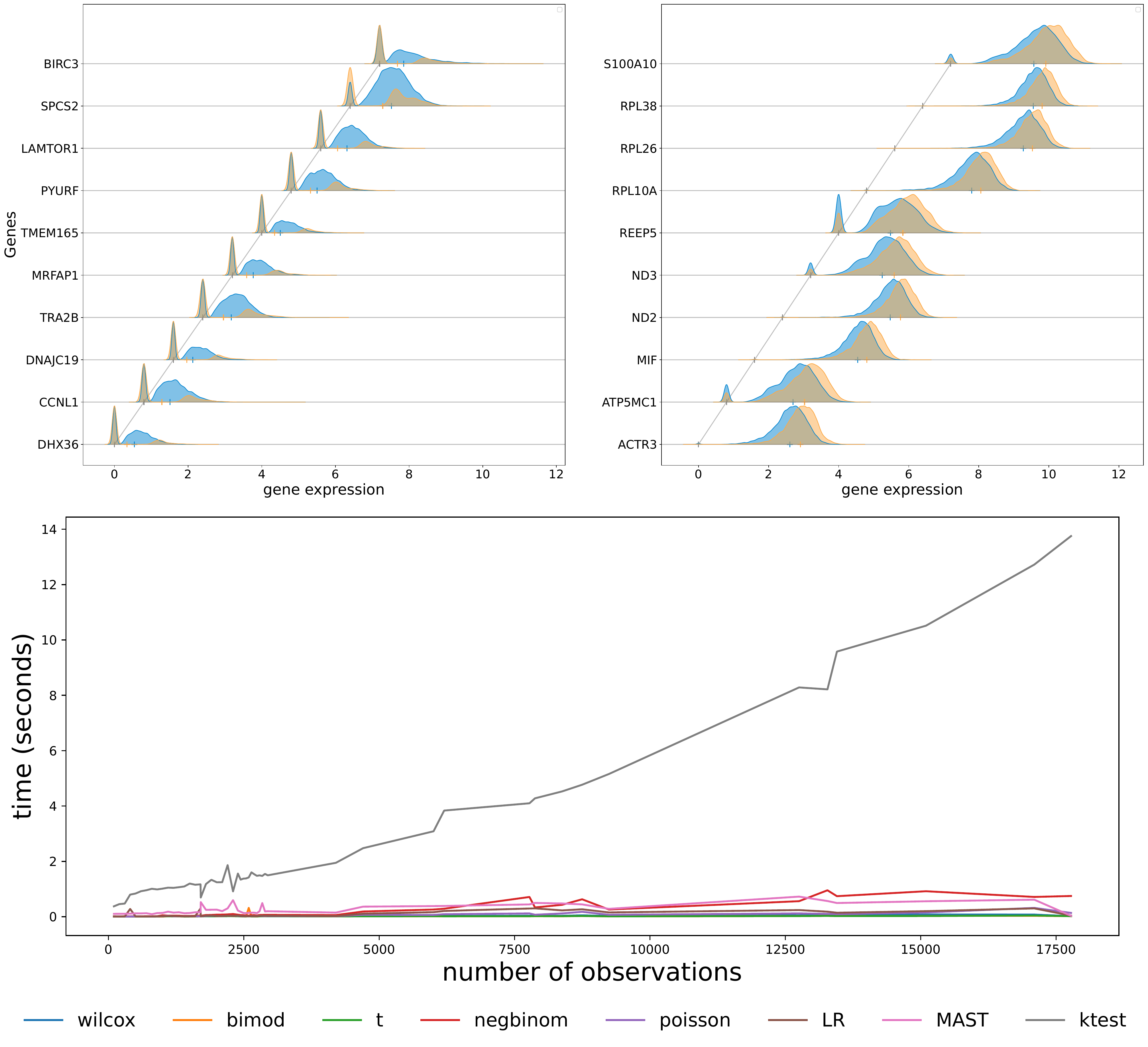} 
\end{center}
    \caption{\revised{Top: Expression densities of the two compared conditions for genes considered as DE by \texttt{ktest}-ZI-kernel and the other single-cell DE methods and considered as non-DE by pseudo-bulk methods. Left: stimulated memory Th0 cells (blue, 4766 cells) vs control memory Th0 cells (orange, 3110 cells) from \cite{cano-gamez_single-cell_2020}.  Right : pig cells stimulated with  lipopolysaccharide (blue, 6605 cells) vs control pig cells (orange, 6148 cells) from \cite{hagai_gene_2018}. Bottom: average computing time (in seconds) of different DEA methods to analyse one gene, according to the number of cells in the sample.}}
\label{fig:gene_distrib}
\end{figure}

\begin{figure}
\begin{center}
\includegraphics[scale=0.3]{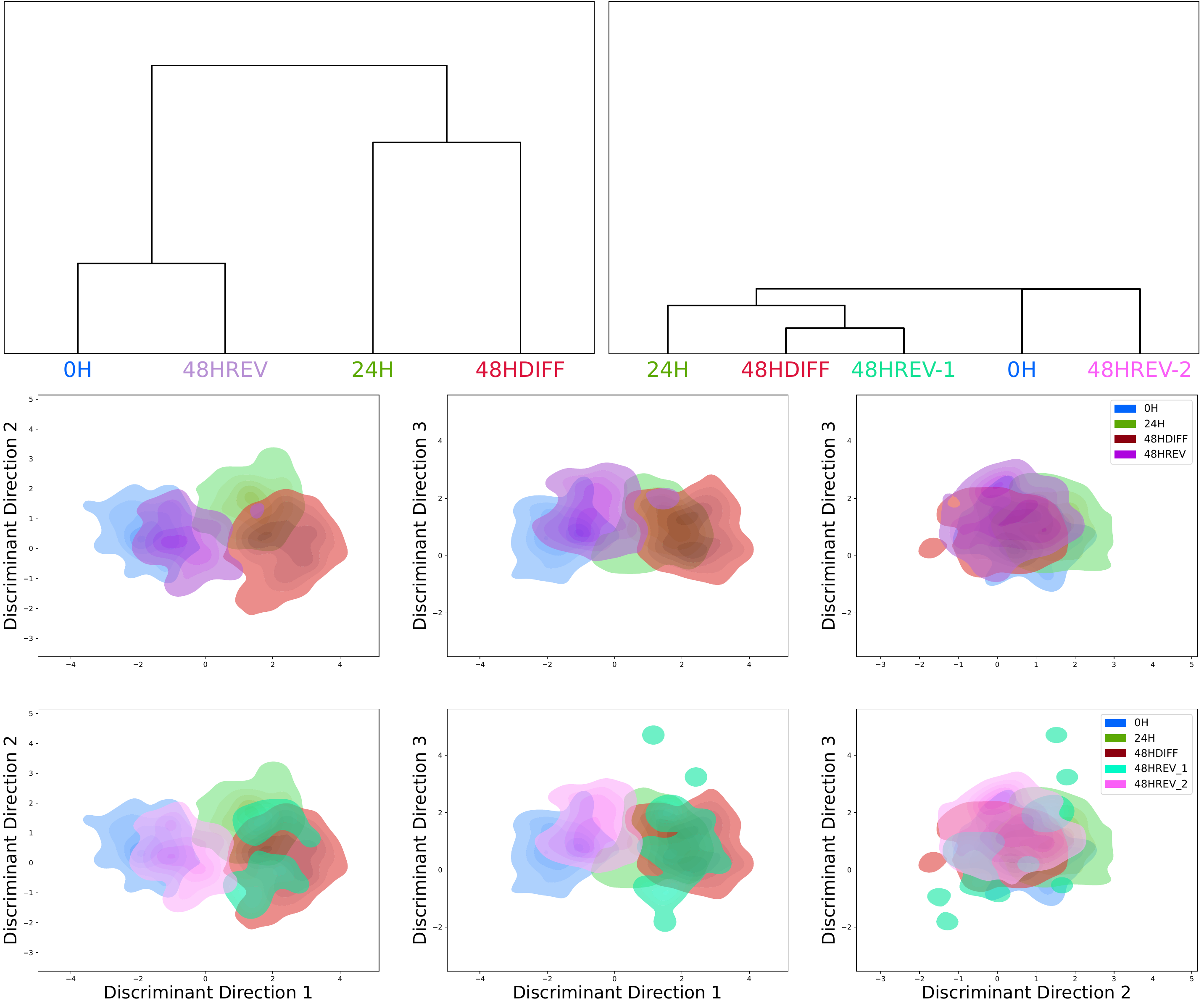}
\end{center}
\caption{Top : Trees from pairwise distances using our test statistic between conditions before (left) and after (right) splitting condition 48HREV into populations 48HREV-1 and 48HREV-2. \revised{Bottom : Cell densities of compared conditions projected on the 3 discriminant axes in the 4-group global comparison  from RTqPCR-Seq data. The multivariate differential expression analysis was performed with $T=10$.}}
\label{fig:reversion RTqPCR multigroupe}
\end{figure}

\begin{figure}
\begin{center}
\includegraphics[scale=0.3]{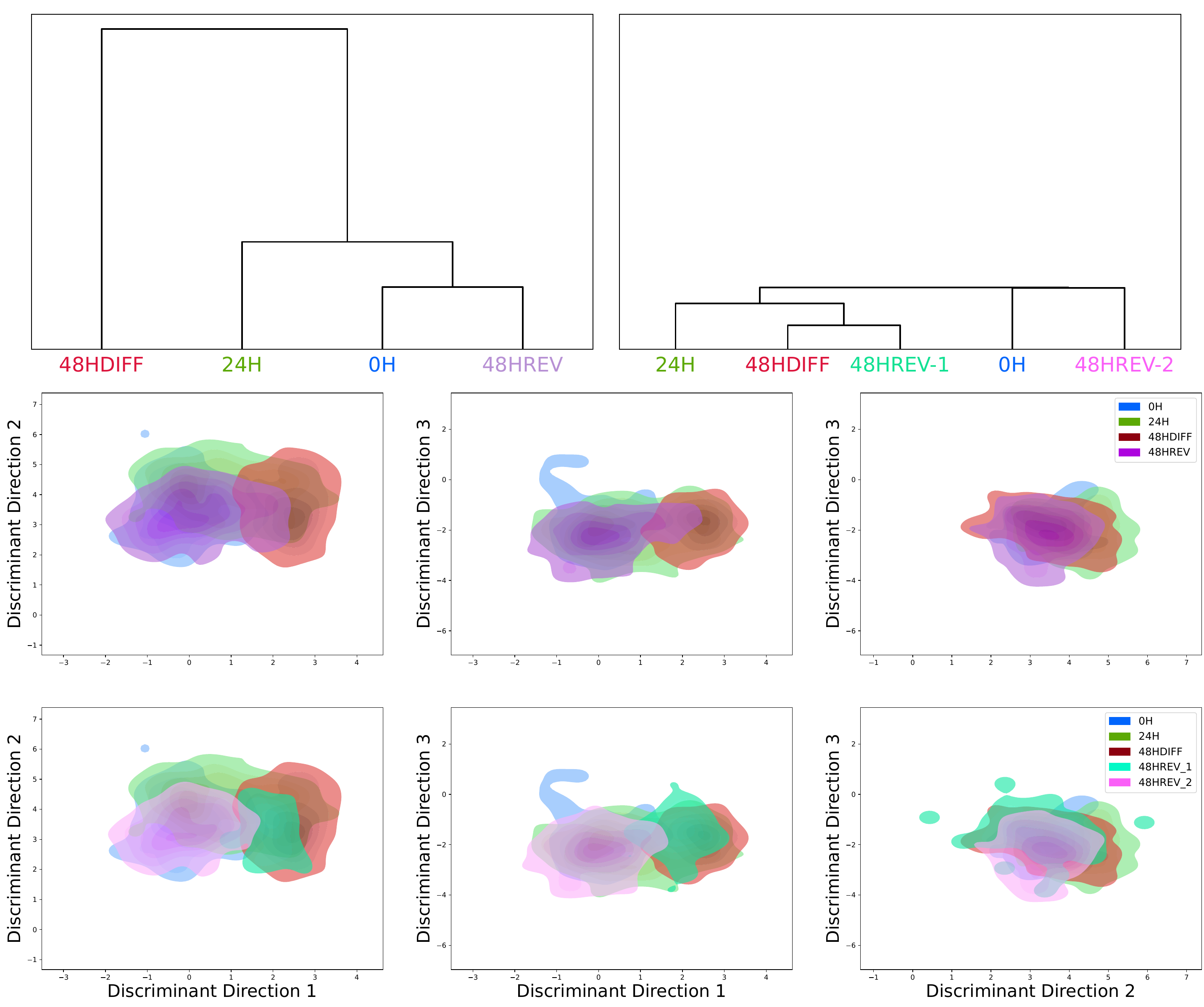}
\end{center}
\caption{Top : Trees from pairwise distances using our test statistic between conditions before (left) and after (right) splitting condition 48HREV into populations 48HREV-1 and 48HREV-2. \revised{Bottom : Cell densities of compared conditions projected on the 3 discriminant axes in the 4-group global comparison  from scRNA-Seq data. The multivariate differential expression analysis was performed with $T=10$.}}
\label{fig:reversion scRNAseq multigroupe}
\end{figure}

\begin{figure}
\begin{center}
\includegraphics[scale=0.3]{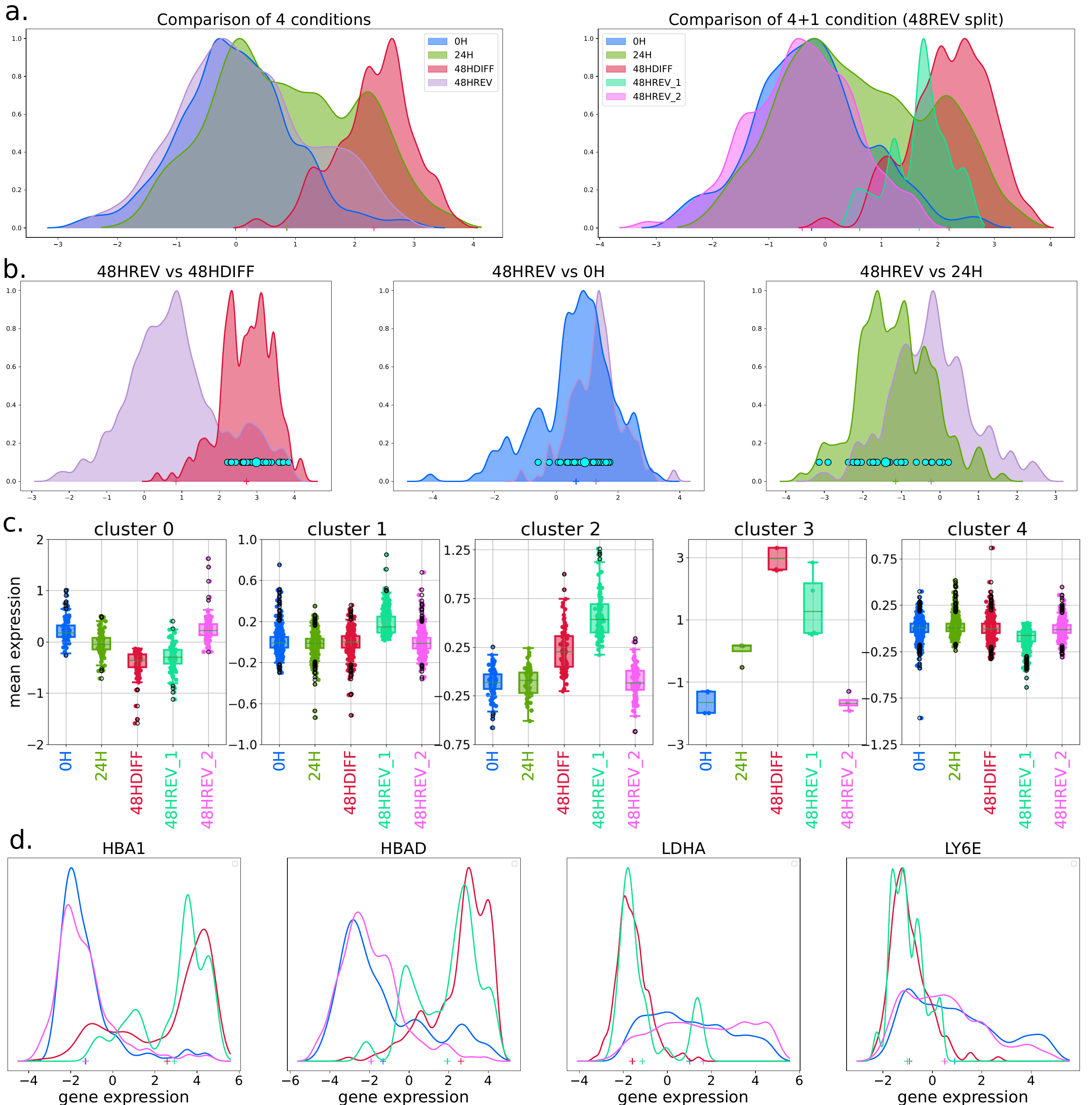}
\end{center}
\caption{a : Cell densities of compared conditions projected on the discriminant axis between conditions 48HREV and 48HDIFF (left), 48HREV and 0H (middle) and 48HREV and 24H (right)  with highlighted population 48HREV-1. c: \revised{Boxplots of the variation of the gene expression along the five populations 0H, 24H, 48HDIFF, 48HREV-1 and 48HREV-2 for the five identified genes clusters.} d : Examples of gene expression distributions in populations 48HREV-1 (turquoise) and 48HREV-2 (pink) compared to populations 0H (blue) and 48HDIFF (red). a,b,c and d are obtained from scRNA-Seq data. \revised{The multivariate differential expression analysis was performed with $T=10$.}}
\label{fig:reversion scRNAseq}
\end{figure}

\begin{figure}
\begin{center}
\includegraphics[scale=0.26]{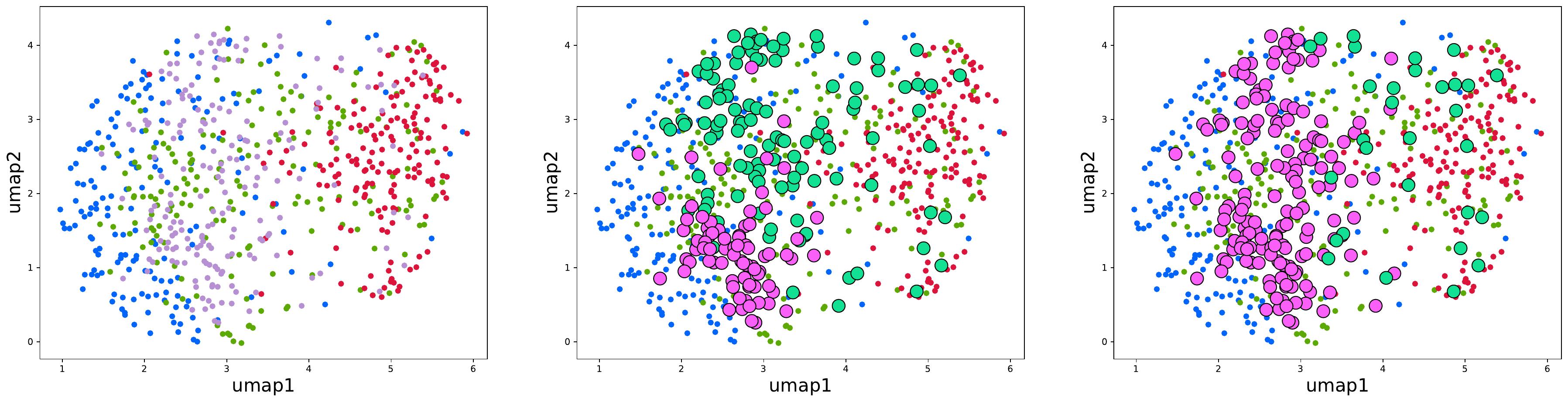}
\end{center}
\caption{Left: Umap representation of the four conditions from scRNA-Seq data (0H (blue), 24H (green) 48HDIFF (red) and 48HREV (purple)). Middle : highlight of the 2 groups of 48HREV identified through a k-means algorithm. Right : The two groups 48HREV-1 (turquoise) and 48HREV-2 (pink) identified on the discriminant axis associated to the truncation parameter $T=10$.}
\label{fig:reversion justification}
\end{figure}

\begin{table}[]
    \centering
    \begin{tabular}{|l|l|}
    \hline
        Method & Description \\
        \hline
        \texttt{ccdf} & DEA based on conditional cumulative distribution function  \cite{gauthier_distribution-free_2021} \\
        \texttt{MAST} & Bayesian logistic regression method  \cite{finak_mast_2015-1} \\
        \texttt{scDD} & Model-based clustering and Bayesian model selection  \cite{korthauer_statistical_2016} \\
        \texttt{SigEMD} & Wasserstein distance between histograms  \cite{wang_sigemd_2018} \\
        \texttt{DEsingle} & Zero-Inflated Negative Binomial regression and likelihood testing  \cite{miao_desingle_2018} \\
        \texttt{SCDE} & Bayesian Negative Binomial modeling and posterior testing  \cite{kharchenko_bayesian_2014} \\
        \hline
    \end{tabular}
\caption{DEA methods compared in the simulation study (Fig. \ref{fig:data_ccdf}).}
\label{tab:methods_simu}
\end{table}

\begin{table}[]
    \centering
    \begin{tabular}{|l|l|}
    \hline
        Single-Cell Methods (Seurat, \cite{butler_integrating_2018}) & Description \\
        \hline
        \texttt{t-test} & Student test \cite{butler_integrating_2018}\\
        \texttt{Wilcoxon} & rank-sum test \cite{butler_integrating_2018}\\ 
        logistic regression & classification based on logistic regression \cite{pmid30664774} \\
        Poisson/ Negative Binomial GLMs & generalized linear models \cite{butler_integrating_2018}  \\
        Mixture Zero-inflation / LogNormal &Likelihood ratio test  \cite{pmid23267174} \\
        Bayesian logistic regression method  & two-part hurdle model (\texttt{MAST}) \cite{finak_mast_2015-1} \\
        \hline
        \hline
        Pseudo Bulk Methods & Description\\
        \hline
        \texttt{DESeq2} & Wald and Likelihood Ratio Test \cite{love_moderated_2014}\\
        \texttt{edgeR} & Likelihood Ratio Test, Quasi Likelihood F-test \cite{robinson_edger_2010}\\
        \texttt{limma} & mean-variance trend considered at the gene level (trend) \\
        & or at the level of each observation (voom) \cite{ritchie_limma_2015}\\
        \hline
    \end{tabular}
\caption{DEA methods compared on the sc-RNASeq datasets \cite{squair_confronting_2021} (Fig. \ref{fig:squair}).}
\label{tab:methods_squair}
\end{table}

\begin{table}[]
    \centering
    \begin{small}
        \begin{tabular}{|lrrccccl|}
        \hline
        chr & start & end & $\widehat{D}_T^2$ &  average  & average  & average  & gene \\ 
         &  &  &   & Persist.   & Persist.-Like & log2FC  &\\
  \hline
chr9 & 133489001 & 133751000 & 123.60  & 0.77 & 1.25 & -0.51 & ADAMTSL2, DBH, SARDH \\ 
  chr5 & 1832001 & 2740000 & 104.70 & 2.45 & 2.80 & -0.43 & IRX4 \\ 
  chr9 & 135509001 & 135802000 & 79.40  & 0.83 & 1.22 & -0.43 & PAEP, LCN1, OBP2A, \\
       &           &          &              &      &     &       & SOHLH1, KCNT1, LCN9 \\ 
  chr9 & 134445001 & 135458000 & 72.50  & 1.67 & 2.12 & -0.49 & OLFM1, FCN2, FCN1, \\
         &           &          &              &      &     &       & COL5A1 \\ 
  chr9 & 76400001 & 77173000 & 51.00  & 1.21 & 1.59 & -0.41 & GCNT1 \\ 
  chr14 & 23331001 & 23355000 & 50.20  & 0.05 & 0.12 & -0.09 & SLC22A17 \\ 
  chr9 & 136123001 & 136206000 & 49.10  & 0.22 & 0.42 & -0.22 & LHX3 \\ 
  chr12 & 129982001 & 130786000 & 48.10  & 1.05 & 1.41 & -0.37 & RIMBP2, PIWIL1 \\ 
  chr3 & 123287001 & 123483000 & 45.20  & 0.69 & 0.91 & -0.26 & ADCY5 \\ 
chr22 & 47950001 & 49760000 & 43.30 & 2.48 & 2.78 & -0.38 & FAM19A5 \\ 

\hline 
\hline
 chr & start & end & $\widehat{D}_T^2$ &  average  & average  & average  & gene \\ 
         &  &    & & Persist.-Like   & Interm. & log2FC  &\\
chr9 & 76400001 & 77173000 & 151.10  & 1.59 & 0.93 & -0.64 & GCNT1 \\ 
  chr11 & 44629001 & 44924000 & 107.70  & 1.19 & 0.65 & -0.50 & TSPAN18 \\ 
  chr1 & 71395001 & 73235000 & 102.10  & 1.94 & 1.37 & -0.51 & NEGR1 \\ 
  chr2 & 120687001 & 121071000 & 99.80  & 1.31 & 0.76 & -0.48 & GLI2 \\ 
  chr3 & 123287001 & 123483000 & 94.80  & 0.91 & 0.47 & -0.43 & ADCY5 \\ 
  chr6 & 33863001 & 34155000 & 90.30 & 1.24 & 0.72 & -0.48 & GRM4 \\ 
  chr11 & 43933001 & 44065000 & 79.80  & 1.04 & 0.61 & -0.40 & ACCSL \\ 
  chr9 & 133489001 & 133751000 & 78.50  & 1.25 & 0.79 & -0.41 & SARDH, DBH, ADAMTSL2 \\ 
  chr5 & 7615001 & 7856000 & 78.20  & 1.20 & 0.69 & -0.42 & C5orf49 \\ 
  chr1 & 44365001 & 44624000 & 73.70  & 1.20 & 0.77 & -0.40 & RNF220 \\ 
\hline
\hline
chr & start & end & $\widehat{D}_T^2$ & average  & average  & average  & gene \\ 
         &    &  & & Interm.   & Naive & log2FC  &\\
chr16 & 85810001 & 86527000 & 466.50  & 2.52 & 2.95 & 0.50 & FOXF1-IRF8 \\ 
chr17 & 73382001 & 74743000 & 337.20  & 3.03 & 3.38 & 0.41 & GPRC5C,CD300A,TTYH2,\\
      &           &          &              &      &     &      & DNAI2, SDK2, RPL38,\\
      &           &          &              &      &     &      & GPR142, CD300C, CD300LD,\\
      &           &          &              &      &     &      & CD300LB, RAB37, KIF19,\\
      &           &          &              &      &     &      & BTBD17, CD300LF, CD300E \\ 
  chr2 & 236192001 & 237072000 & 314.50  & 1.99 & 2.41 & 0.48 & IQCA1,ASB18 \\ 
  chr1 & 195633001 & 196663000 & 249.90 & 0.92 & 0.47 & -0.51 & KCNT2,CFH \\ 
  chr1 & 189337001 & 190666000 & 235.80  & 1.03 & 0.58 & -0.49 & BRINP3 \\ 
  chr11 & 97578001 & 99922000 & 229.30  & 1.65 & 1.18 & -0.55 & CNTN5 \\ 
  chr1 & 215645001 & 217423000 & 221.10  & 1.79 & 1.30 & -0.59 & ESRRG,USH2A \\ 
  chr1 & 71395001 & 73235000 & 215.20  & 1.37 & 0.91 & -0.52 & NEGR1 \\ 
  chr20 & 59052001 & 59846000 & 213.30  & 2.05 & 2.36 & 0.35 & EDN3,PHACTR3 \\ 
  chr7 & 14654001 & 15126000 & 209.30  & 0.70 & 0.34 & -0.37 & DGKB \\ 
\hline
\hline
chr & start & end & $\widehat{D}_T^2$ &  average  & average  & average  & gene \\ 
         &    &  & & Persist.-Like   & Naive & log2FC  &\\
chr1 & 71395001 & 73235000 & 292.20  & 1.94 & 0.91 & -1.05 & NEGR1 \\ 
  chr7 & 15136001 & 16070000 & 250.40  & 1.65 & 0.70 & -0.95 & MEOX2,AGMO \\ 
  chr2 & 192142001 & 193818000 & 237.00 & 1.64 & 0.74 & -0.90 & TMEFF2 \\ 
  chr11 & 97578001 & 99922000 & 230.30  & 2.04 & 1.18 & -0.86 & CNTN5 \\ 
  chr11 & 44629001 & 44924000 & 217.10  & 1.19 & 0.39 & -0.77 & TSPAN18 \\ 
  chr7 & 14654001 & 15126000 & 203.80  & 1.09 & 0.34 & -0.72 & DGKB \\ 
  chr1 & 195633001 & 196663000 & 201.50  & 1.31 & 0.47 & -0.83 & KCNT2,CFH \\ 
  chr9 & 76400001 & 77173000 & 199.90  & 1.59 & 0.68 & -0.91 & GCNT1 \\ 
  chr16 & 85810001 & 86527000 & 199.30  & 2.29 & 2.95 & 0.93 & FOXF1,IRF8 \\ 
  chr5 & 7615001 & 7856000 & 188.20 & 1.20 & 0.43 & -0.73 & C5orf49 \\ 
  \hline
\end{tabular}
    \end{small}
    \caption{Differential analysis of sc-chIPseq data: top-10 differential regions for pairwise comparisons between persister cells and the three sub-populations of untreated cells. Adjusted $p$-values are $<10^{-3}$ (Bonferroni correction). The last Gene column corresponds to the genes overlapping the regions.}
    \label{tab:chip}
\end{table}

\newpage
\bibliographystyle{abbrv}
\bibliography{OLFD24}

\end{document}